\documentclass[10pt,twocolumn,letterpaper]{article}

\usepackage{cvpr}
\usepackage{times}
\usepackage{epsfig}
\usepackage{graphicx}
\usepackage{amsmath}
\usepackage{amssymb}

\usepackage{mathtools}
\usepackage{algorithm}
\usepackage{algorithmic}
\usepackage{nicefrac}

\DeclarePairedDelimiter\round{\lceil}{\rfloor}

\newcommand{\N}{{\mathbb N}}
\newcommand{\R}{{\mathbb R}}
\newcommand{\M}[1]{\mathtt{#1}}
\newcommand{\V}[1]{\mathbf{#1}}

\def\gb{Gr{\"o}bner basis\xspace}

\def\squad{\hskip0.55em\relax}
\usepackage[pagebackref=true,breaklinks=true,letterpaper=true,colorlinks,bookmarks=false]{hyperref}

 \cvprfinalcopy 

\def\cvprPaperID{2003} 

\ifcvprfinal\fi
\begin{document}

\title{On the Two-View Geometry of Unsynchronized Cameras}

\author{%
Cenek Albl$^\textrm{1}$ $\squad$
Zuzana Kukelova$^\textrm{1}$ $\squad$
Andrew Fitzgibbon$^\textrm{2}$ $\squad$
Jan Heller $^\textrm{3}$ $\squad$
Matej Smid$^\textrm{1}$ $\squad$
Tomas Pajdla$^\textrm{1}$
\and
$^\textrm{1}$Czech Technical University in Prague\\
Prague, 
Czechia\\
{\tt\small \{alblcene,kukelova\}@cmp.felk.cvut.cz}\\
{\tt\small smidm@cmp.felk.cvut.cz,pajdla@cvut.cz}\\
\and
$\!\!^\textrm{2}$Microsoft \\  Cambridge,
UK\\
{\tt\small awf@microsoft.com}
\and
$^\textrm{3}$Magik Eye Inc.\\
New York,
US\\
{\tt\small jan@magik-eye.com}
}


\maketitle
\begin{abstract}
   \noindent We present new methods for simultaneously estimating camera geometry and time shift from video sequences from multiple unsynchronized cameras.  Algorithms for simultaneous computation of a fundamental matrix or a homography with unknown time shift between images are developed. Our methods use minimal correspondence sets  (eight for fundamental matrix and four and a half for homography) and therefore are suitable for robust estimation using RANSAC. Furthermore, we present an iterative algorithm that extends the applicability on sequences which are significantly unsynchronized, finding the correct time shift up to several seconds. We evaluated the methods on synthetic and wide range of real world datasets and the results show a broad applicability to the problem of camera synchronization.
\end{abstract}
\section{Introduction}
\noindent Many computer vision applications, e.g., human body modelling~\cite{Muhavi2010, deutscher2005articulated}, person tracking~\cite{Ellis,Zhou2016}, pose estimation~\cite{frahm2004pose}, robot navigation~\cite{carrera2011lightweight,Kitti}, and 3D object scanning~\cite{seitz2006comparison}, benefit from using multiple-camera systems. In tightly-controlled laboratory setups, it is possible to have all cameras temporally synchronized. However, applicability of multi-camera systems could be greatly enlarged when cameras might run without synchronization~\cite{hasler2009markerless}. Synchronization is sometimes not possible, e.g.\ in automotive industry, but even if it was possible, using asynchronous cameras may produce other benefits, e.g., reducing bandwidth requirements and improving temporal resolution of event detection and motion recovery~\cite{Elhayek2012}.

In this paper, we (1) introduce {\em practical solvers} that simultaneously compute either a fundamental matrix or a homography and time shift between image sequences, and (2) we propose a fast {\em iterative algorithm} that uses RANSAC~\cite{Fischler1981} with our solvers in the inner loop to synchronize large time offsets. {\em Our approach can accurately calibrate large time shifts, which was not possible before}. 

\begin{figure}[t]
\centering
\includegraphics[width=0.8\linewidth]{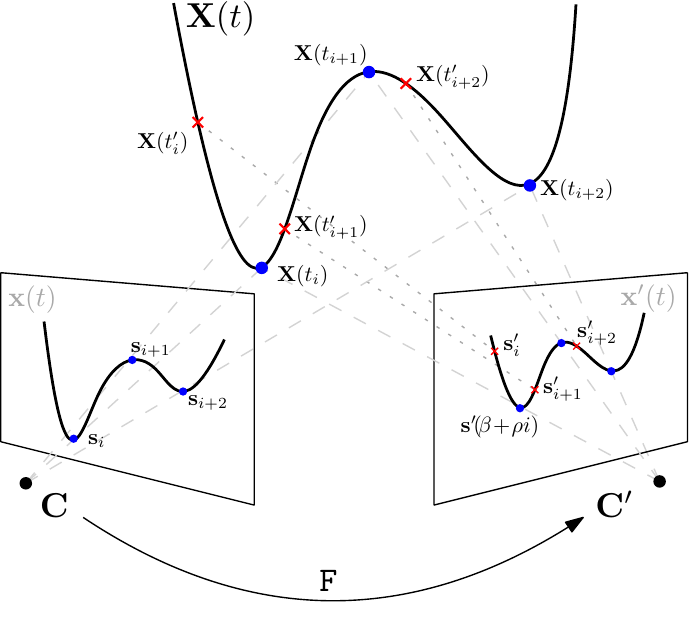}
\caption{Two cameras capture a moving point at different times, so the projection rays of the two cameras meet nowhere.}
\label{fig:teaser}
\end{figure}
\subsection{Related work}
\noindent Many video and/or image sequence synchronization methods are based on image content analysis~\cite{Pundik2010,Caspi2002,Tuytelaars2004,Dai2006,Lei2006,Elhayek2012a,Padua2010,Rao2003,Stein1999}, or on synchronizing video by audio tracks~\cite{Shrestha2007} and therefore their applicability is limited. Other approaches employed compressed video bitrate profiles~\cite{Schroth2010} and still camera flashes~\cite{Shrestha2006}.The methods differ in temporal transformation models. Often, time shift~\cite{Pundik2010, Stein1999, Tuytelaars2004, Lei2006}, or time shift combined with variable frame rate 
~\cite{Elhayek2012a, Padua2010, Caspi2002}, are used.
The majority of previous work requires rigid sets of cameras. Notable examples of synchronization methods for independently moving cameras are~\cite{Tuytelaars2004, Lei2006}.

Many methods share a similar basis. A set of trajectories is detected in every video sequence using an interest point detector and an association rule or a 2D tracker. The trajectories are matched across sequences. A RANSAC based algorithm is often used to estimate jointly or in an iterative manner the parameters of temporal and spatial transformations~\cite{Elhayek2012a, Padua2010, Caspi2002}. In~\cite{Elhayek2012a}, RANSAC is used to search for matching trajectory pairs in filtered set of all combinations of trajectories in a sequence pair. The epipolar geometry has to be provided. The method \cite{Padua2010} enables joint synchronization of $N$ sequences by fitting a single $N$-dimensional line called {\em timeline} in a RANSAC framework. The algorithm \cite{Caspi2002} estimates temporal and spatial transformation based on tentative trajectory matches. 

Methods using exhaustive search to find the homography~\cite{Stein1999} and either fundamental matrix or homography~\cite{Tresadern2009} along with the time offset were presented. These are searching over the entire space of possible time shifts.

The two most closely related works to ours are~\cite{Noguchi2007,Nischt2009} that jointly estimate two-view geometry together with time shift from approximated image point trajectories. In~\cite{Noguchi2007} estimated epipolar geometry or homography along with time shift using non-linear least squares, approximating the image trajectory by a straight line. The algorithm is initialized by the 7pt algorithm~\cite{hartley2003} and a zero time shift. Work~\cite{Nischt2009} extended this approach by estimating difference in frame rate and using splines instead of lines. Both the above works achieve good results only when given a good initialization, e.g., on sequences less than $0.5$ seconds time shift and with no gross matching errors.

\subsection{Contribution}
\noindent In this paper we present two new contributions. 

First, we present a new method for {\em simultaneous computation of two-view camera geometry and temporal offset parameters from minimal sets of point correspondences}. We solve for fundamental matrix or homography together with temporal offset of image sequences. Our methods need only moving image point trajectories, which are easy to track. Unlike in~\cite{Noguchi2007,Nischt2009}, we use a small (minimal) numbers of correspondences and we therefore are robust to outliers when combined with RANSAC robust estimation. 

Secondly, we present an {\em iterative scheme using the minimal solvers to efficiently estimate large time offsets}. Our approach is based on RANSAC loop running our minimal solvers. This approach efficiently searches in the space of possible time offsets, which is much more efficient than exhaustive search methods~\cite{Stein1999,Tresadern2009} developed before. 

We evaluated our approach on a wide range of scenes and demonstrated its capability of synchronizing various kinds of real camera setups, such as driving cars, surveillance cameras, or sports match recordings with no other information than image data.

We demonstrate that our solvers are able to synchronize small time shifts of fractions of a second as well as large time shifts of tens of seconds. Our iterative algorithm is capable of synchronizing medium time shifts (i.e.\ tens of frames) with less than 5 RANSAC iterations and large time offsets (i.e.\ tens to hundreds of frames) using tens of RANSAC iterations. Overall, our approach is much more efficient than other methods utilizing RANSAC~\cite{Padua2010}. 

By solving two-camera synchronization problem, we also solve the multi-camera synchronization problem since temporal offsets of multiple cameras can be determined pairwise to serve as the initialization point for a global iterative solutions based on bundle adjustment~\cite{Triggs1999}.
%
%
\section{Problem formulation}
\noindent
Let us consider two unsynchronized cameras with a fixed relative pose~\cite{hartley2003} producing a stereo video sequence by observing a dynamic scene. Motions of objects in the video sequence are indistinguishable from  camera rig motions, and therefore, we will present the problem for static cameras and moving objects.
\subsection{Geometry of two unsynchronized cameras}
\noindent
The coordinates of a 3D point moving along a smooth trajectory in space can be described by function
\begin{eqnarray}
\V{X}(t)=[X_1(t),X_2(t),X_3(t),1]^\top,
\end{eqnarray}
where $t$ denotes time, see Figure~\ref{fig:teaser}. Projecting $\V{X}(t)$ into the image planes of the two distinct cameras produces two 2D trajectories $\V{x}(t)$ and $\V{x}'(t)$.  Now, let's assume that the first camera captures frames with frequency $f$ (period $p = \nicefrac{1}{f}$) starting at time $t_0$. This leads to a sequence of samples 
\begin{eqnarray}
\V{s}_i = [u_i,v_i,1]^\top = \V{x}(t_i) = \pi(\V{X}(t_i)), \, i = 1,\ldots,n.
\end{eqnarray}
of the trajectory $\V{x}(t)$ at times $t_i = t_0 + i p$. 

Analogously, assuming a sampling frequency $f'$ (period $p'=\nicefrac{1}{f'}$), at times $t'_j = t'_0+ j p'$, the second camera produces a sequence of samples
\begin{eqnarray}
\V{s}'_j=[u'_j,v'_j,1]^\top = \V{x}'(t'_j) = \pi^\prime(\V{X}(t^\prime_j)),\, j = 1,\ldots,n'.
\end{eqnarray}

In general, there is no correspondence between the $s_i$ and $s'_j$ samples, \ie, for $i=j$, $s_i$ and $s'_j$ do not represent the projections of the same 3D point. There are two main sources of desynchronization in video streams. The first one is the different recording starts or camera shutters triggering independently leading to a constant time shift. The second source are different frame rates or imprecise clocks leading to different time scales. Assuming these two sources, we can map the time $t$ to $t^\prime$ for frame $i$ using $\j(i): \N \rightarrow \R$ as
%
\begin{eqnarray}
\j(i) = \frac{\scriptstyle t_i-t'_0}{\scriptstyle p'} = \frac{\scriptstyle t_0 + ip - t'_0}{ \scriptstyle p'}  = \frac{\scriptstyle t_0 - t'_0}{\scriptstyle p'} + \frac{\scriptstyle p}{\scriptstyle p^\prime}i = \beta + \rho i,
\label{eq:dsmodel}
\end{eqnarray}
where $\beta\in\mathbb{R}$ is captures the time shift and $\rho\in\mathbb{R}$ the time scaling. Note that $\j(i)$ is an integer-to-real linear mapping with an analogous inverse mapping $\i(j)$. 
Given the model in (\ref{eq:dsmodel}) and a sequence of image samples $\V{s}'_j,j=1,\ldots,n'$, we can interpolate a continuous curve $\V{s}'(\j)$, for example using a spline, so that the 2D point corresponding to $\V s_i$ is approximately given as
\begin{eqnarray}
\V s_i \longleftrightarrow \V s'(\beta + \rho i).
\end{eqnarray}
Notice that the interpolated image curve $\V s'(\cdot)$ is not equivalent to the true image trajectory $\V x'(\cdot)$, but may be expected to be a good approximation 
under certain conditions.
Even though it might appear reasonable to assume time shift to be known within a fraction of a second, it is often the case in practice that the timestamps are based on CPU clocks which together with startup delays can lead to time shift $\beta$ being in the order of seconds. On the other hand, the time scaling $\rho$ is more often known or can be calculated accurately.

\subsection{Epipolar geometry}
\noindent
At any given (real-valued) time $t$, the epipolar constraint of the two cameras is determined by the following equation:
\begin{equation}
    \V{x}'(t)^{\top} \M{F} \V{x}(t) = 0.
    \label{eq:epipolar}
\end{equation}
%
%
For a sample $\V s_i$ in the first camera, we can rewrite (\ref{eq:epipolar}) using the corresponding point $\V x'(t_i)$ in the second camera as
\begin{equation}
    \V{x}'(t_i)^{\top} \M{F} \V{s}_i = 0,
    \label{eq:timediffts}
\end{equation}
%
Using the approximation of the trajectory $\V x'$ by $\V s'$, we can express the approximate epipolar constraint as
\begin{equation}
    \V{s}'(\beta + \rho i)^{\top} \M{F} \V{s}_i = 0.
    \label{eq:timedifftss}
\end{equation}
In principle, we can solve for the unknowns $\beta, \rho$, and $\M F$ given 9 correspondences $s_i, s'_j$.  However, such a solution would be necessarily iterative and too slow to be used as a RANSAC kernel. In the following, a further approximation is used to expresses the problem as a system of polynomials, which can be solved efficiently~\cite{Kukelova-ECCV-2008}.  In \S\ref{sec:iter} we show an iterative solution built on this kernel, which can recover offsets of up to hundreds of frames.

\subsection{Linearization of $\V{s}^{\prime}$ for known $\rho$}
Let us assume that the relative framerate $\rho$ is known.
In practice, the image curve $\V s^{\prime}$ is a complicated object.
To arrive to our polynomial solution we approximate $\V s^{\prime}$ by the first order Taylor polynomial at $\beta_0 + \rho i$  
\begin{equation}
\V{s}'(\beta + \rho i) \approx \V{s}'(\beta_0 + \rho i) + (\beta - \beta_0) \V v = \V s^{\prime\prime}(\beta + \rho i) 
\label{eq:suv}
\end{equation}
where $\V v$  is the tangent vector $\dot{\V s}'(\beta_0 + \rho i)$, and   $\beta_0$ is an initial time shift estimate.
We denote this approximation as $\V s^{\prime\prime}$.

Further, we choose $\V{v}$ to approximate the tangent over the next $d$ samples.   Let $j_0 = \lfloor\beta_0 + \rho i\rfloor$ be the approximate discrete correspondence, and then
\begin{equation}
   \V{v} = \V{s}'_{j_0+d}-\V{s}'_{j_0}.
   \label{eq:v}
\end{equation}
Note, that now $\V{v}$ depends on $i$.
For compactness, we write $\V u_i = \V{s}'(\beta_0 + \rho i) - \beta_0 \V v_i $, and (\ref{eq:timedifftss}) becomes
\begin{equation}
    (\V u_i + \beta \V v_i)^{\top} \M{F} \V{s}_i = 0
    \label{eq:shift_epipolar}
\end{equation}


\begin{figure}[t]
\includegraphics[width=0.49\linewidth]{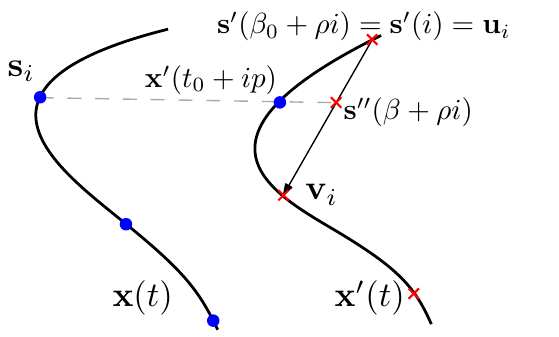}
\includegraphics[width=0.49\linewidth]{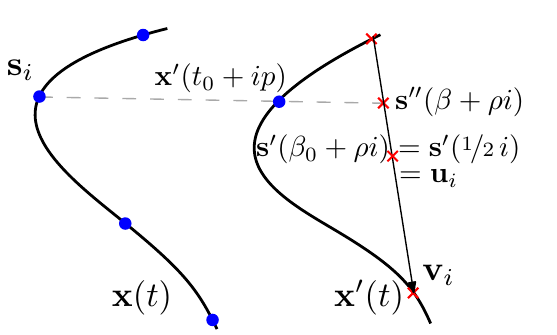}
\caption{Illustration of the proposed trajectory linearization. (Left) Situation for $\rho = 1, \, \beta_0 = 0$ and $d=1$ (Right) Situation for $\rho = 1/2, \, \beta_0 = 0$ and $d=1$.}
\label{fig:illustration}
\end{figure}


 In the rest of the paper, we will assume that $f = f'$ and the initial estimate $\beta_0 = 0$. This situation is illustrated in Figure~\ref{fig:illustration} (Left). 
 However the key results  hold for general known $\rho$,  Figure~\ref{fig:illustration} (Right),  and $\beta_0 \neq 0$.



\subsection{Homography}
Using the same approach, we can write the equation for homography between two unsynchronized cameras. In the synchronized case, the homography between two cameras can be expressed as
\begin{equation}
    \M{H}\V{s}_i = \lambda_i\V{s}'_i.
    \label{eq:homo}
\end{equation}
where $\lambda_i$ is an unknown scalar. Approximating the image motion locally by a straight line gives for two unsynchronized cameras
\begin{equation}
    \M{H}\V{s}_i = \lambda_i\left(\V{u}_i+\beta \V{v}_i \right).
    \label{eq:shift_homo}
\end{equation}

\section{Solving the equations}
\label{sec:solver}

\subsection{Minimal solution to epipolar geometry}
\label{sec:minimalF}
The minimal solution to the simultaneous estimation of the epipolar geometry and the unknown time shift $\beta$ starts with the epipolar constraint~\eqref{eq:shift_epipolar}.
The fundamental matrix $\M{F} = \left[f_{ij}\right]_{i,j=1}^3$ is a $3 \times 3$  singular matrix, \ie it satisfies
\begin{equation}
    \det(\M{F}) = 0.
    \label{eq:rank}
\end{equation}
Therefore, the minimal number of samples $\V{s}_i$ and $\V{s}'_i$ necessary to solve this problem is eight.

For eight samples in general position in two cameras, the epipolar constraint~\eqref{eq:shift_epipolar} can be rewritten as
\begin{equation}
    \M{M}\V{w} = \V{0},
\end{equation}
where $\M{M}$ is a $8 \times 15$ coefficient matrix of rank 8 and $\V{w}$ is a vector of monomials $\V{w} = [f_{11}, f_{12}, f_{13}, f_{21},$ $ f_{22}, f_{23}, f_{31}, f_{32}, f_{33}, \beta f_{11}, \beta f_{12}, \beta f_{13}, \beta f_{21},  \beta f_{22}, \beta  f_{23}]$. Since the fundamental matrix is only given up to scale, the monomial vector $\V{w}$ can be parametrized using the 7-dimensional nullspace of the matrix $\M{M}$ as
\begin{equation}
    \V{w} = \V{n}_0 + \textstyle \sum_{i=1}^{6}\alpha_i \V{n}_i,
    \label{eq:u_param}
\end{equation}
where $\alpha_i$, $i=1,\dots, 6$ are new unknowns and $\V{n}_i$, $i=0,\dots, 6$ are the null space vectors of the coefficient matrix $\M{M}$. The elements of the monomial vector $\V{w}$ satisfy
\begin{equation}
    \beta w_j = w_k \quad\text{for}~(j,k)\in\{(1,10),...,(6,15)\}.
    \label{eq:u_const}
\end{equation}

The parametrization \eqref{eq:u_param} used in the rank constraint~\eqref{eq:rank} and in the quadratic constraints~\eqref{eq:u_const} results in a quite complicated system of 7  polynomial equations in 7 unknowns $\alpha_1,\dots,\alpha_6,\beta$. 
Therefore, we first simplify these equations by eliminating the unknown time shift $\beta$ from these equations using the elimination ideal method presented in~\cite{kukelova2017elimination}. This results in a system of 18 equations in 6 unknowns $\alpha_1,\dots,\alpha_6$. Even though this system contains more equations than the original system, its structure is less complicated.
We solve this system using the automatic generator of \gb solvers~\cite{Kukelova-ECCV-2008}. The final \gb solver performs Gauss-Jordan elimination of a $194 \times 210$ matrix and the eigenvalue computations of a $16 \times 16$ matrix, since the problem has $16$ solutions.
Note that by simply applying~\cite{Kukelova-ECCV-2008} to the original system of 7 equations in 7 unknowns  a huge and numerically unstable solver of size $633 \times 649$ is obtained.

\subsection{Generalized eigenvalue solution to epipolar geometry}
\label{sec:GEP}
\noindent Using the non-minimal number of nine point correspondences, the epipolar constraint~\eqref{eq:shift_epipolar} can be rewritten as 
\begin{eqnarray}
    (\M{M}_1 +\beta \M{M}_2) \V{f} = \V{0},
    \label{eq:polyeig}
\end{eqnarray}
where $\M{M}_1$ and $\M{M}_2$ are $9 \times 9$ coefficient matrices and $\V{f}$ is a vector containing nine elements of the fundamental matrix~$\M{F}$

The formulation~\eqref{eq:polyeig} is a generalized eigenvalue problem (GEP) for which efficient
numerical algorithms are readily available. The eigenvalues of~\eqref{eq:polyeig} give us solutions to $\beta$ and the eigenvectors to fundamental matrix~$\M{F}$.

For this problem the rank of the matrix $\M{M}_2$ is only six and three from nine eigenvalues of~\eqref{eq:polyeig} are always zero.
Therefore, instead of $\textrm{9} \times \textrm{9}$ 
we can solve only $\textrm{6} \times \textrm{6}$ 
GEP.

This generalized eigenvalue solution is more efficient than the minimal solution presented in section~\ref{sec:solver}, however note that the GEP solution uses non-minimal number of nine point correspondences and the resulting fundamental matrix does not necessarily satisfy $\det(\M{F})=0$.

\subsection{Minimal solution to homography estimation}
The minimal solution to the simultaneous estimation of the homography and the unknown time shift $\beta$ starts with the equations of the form~\eqref{eq:shift_homo}.

First, the solver eliminates the scalar values $\lambda_i$ from~\eqref{eq:shift_homo}. This is done by
multiplying~\eqref{eq:shift_homo} by  the skew symmetric matrix
$\left[\V{u}_i+\beta\V{v}_i\right]_{\times}$.
This leads to the matrix equation 
\begin{equation}
    \label{eq:homography_eq_skew}
    \left[\V{u}_i+\beta\V{v}_i\right]_{\times}\M{H}\V{s}_i = \V{0}.
\end{equation}
%
%

The matrix equation~\eqref{eq:homography_eq_skew} contains three polynomial
equations from which only two are linearly independent, because the skew symmetric  matrix  has rank two.
This means that we need at least 4.5 (5) samples in two images to estimate the
unknown homography $\M{H}$ as well as the time shift~$\beta$.

Now let us 
use the equations corresponding to the first and second row of the matrix equation~\eqref{eq:homography_eq_skew}. In these equations  $\beta$ multiplies only the $3^{rd}$ row of the unknown homography matrix.
This lead to nine homogeneous   equations   in   12   monomials $\V{w}
=[ h_{11}, h_{12}, $ $ h_{13}, h_{21}, h_{22}, h_{23},h_{31}, h_{32}, h_{33}, \beta\,h_{31}, \beta\,h_{32},\beta\,h_{33}]^{\top}$ for 4.5 samples in two images  (i.e.\ we use only one equation from the three equations~\eqref{eq:homography_eq_skew} for the 5th sample). 

We can stack these nine equations into a matrix form $\M{M}\,\V{w} = \V{0}$,
%
%
where  $\M{M}$ is  a  $\textrm{9} \times  \textrm{12}$  coefficient matrix.
Assuming that $\M{M}$ has  full rank equal to nine, \ie, we have 
non-degenerate samples, the dimension of $\mathrm{null}(\M{M})$
is 3. This means that the monomial vector $\V{w}$ can in general be rewritten
as a linear combination of three null space basis vectors $\V{n}_i$ of the
matrix $\M{M}$ as
\begin{eqnarray}
	\label{eq:null}
	\V{w} = \textstyle    \sum_{i=1}^{3} \gamma_i\,\V{n}_i,
\end{eqnarray}
where $\gamma_i$ are new unknowns. Without loss of generality, we can set $\gamma_3 = 1$ to fix the scale of the homography and to bring down the number of unknowns.  
For 5 or more samples, instead of null space vectors $\V{n}_i$, we use in~\eqref{eq:null} three right singular vectors corresponding to three smallest singular values of $\M{M}$.

The elements of the monomial vector $\V{w}$ are not
independent. We can see that $w_{10} = \beta\,w_7,  w_{11} = \beta\,w_8$, and $w_{12} = \beta\,w_9$,
where $w_i$ is the $i^{th}$ element of the vector $\V{w}$. 
These three constraints, together with the parametrization from equation~(\ref{eq:null}) form a system of three quadratic equations in three unknowns $\gamma_1, \gamma_2$, and $\beta$ and only 6 monomials.
This system of three equations has a very simple structure and can be directly solved by performing G-J elimination of the $3 \times 6$ coefficient matrix $\M{M}_1$ representing these tree polynomials, and then by computing eigenvalues of the $3 \times 3$ matrix obtained from this eliminated matrix  $\M{M}_1$. This problem results in up to three real solutions. 

Note, that the problem of estimating homography and  $\beta$ can also be formulated as a generalized eigenvalue problem, similarly as the problem of estimating epipolar geometry (Section~\ref{sec:GEP}).
However, due to the lack of space and the fact that the presented minimal solution is extremely efficient, we do not describe the GEP homography solution here.
\section{Using RANSAC}
In this section we would like to emphasize the role of RANSAC for our solvers. RANSAC is generally used for robustness since the minimal solvers are sensitive to noise and outliers. Outliers in the data will usually come from two sources. One are the mismatches and misdetections and the other is the non-linearity of the point trajectory. Even without gross outliers due to false detections, there will always be outliers with respect to the model in places where the trajectory is not straight on the interpolating interval. Therefore, it is usually beneficial to use RANSAC even if we are sure the correspondences are precise.

By using RANSAC, we avoid those parts of the trajectory and pick the parts that are approximately straight and linear in velocity. Basically we only need to sample 8(F) or 5(H) parts of the trajectory where this assumption holds to obtain a good model, even if the rest of the trajectory is highly non-linear.

\section{Performance of the solvers on synthetic data}
\label{sec:exp_synthetic}
\noindent
First, we investigated the performance of estimating the time shift $\beta$ using the proposed $\M{F}$ and $\M{H}$ minimal solvers. We simulated a random movement of a 3D point in front of two cameras. The simulated 3D trajectory was then sampled at different times in each camera, the difference being the ground truth time shift $\beta_{gt}$. Image noise was added from a normal distribution with  $\sigma=0.5$~px. We tested the minimal solvers with various interpolation distances $d$ and compared them also to the standard seven point fundamental matrix (7pt-F) and four point homography (4pt-H) solvers~\cite{hartley2003}.  Each algorithm was tested on 100 randomly generated scenes for each $\beta_{gt}$, resulting in tens of thousands of experiments.

\begin{figure}
\includegraphics[width=0.48\columnwidth]{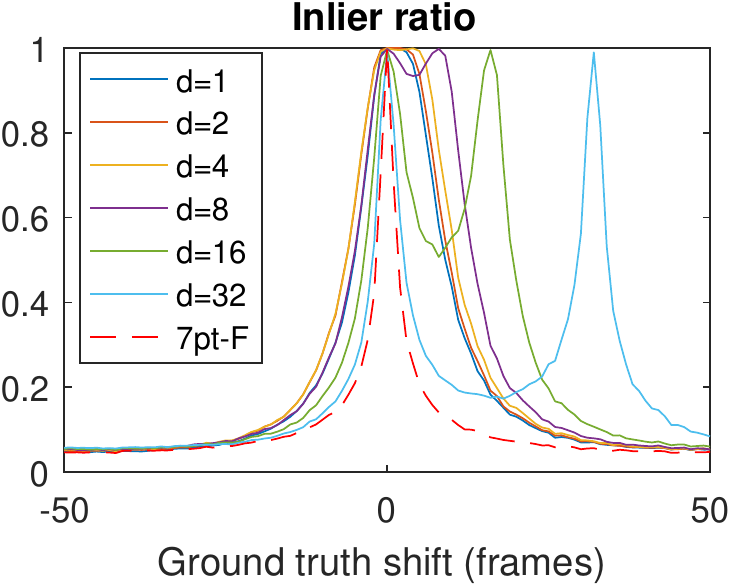}
\includegraphics[width=0.48\columnwidth]{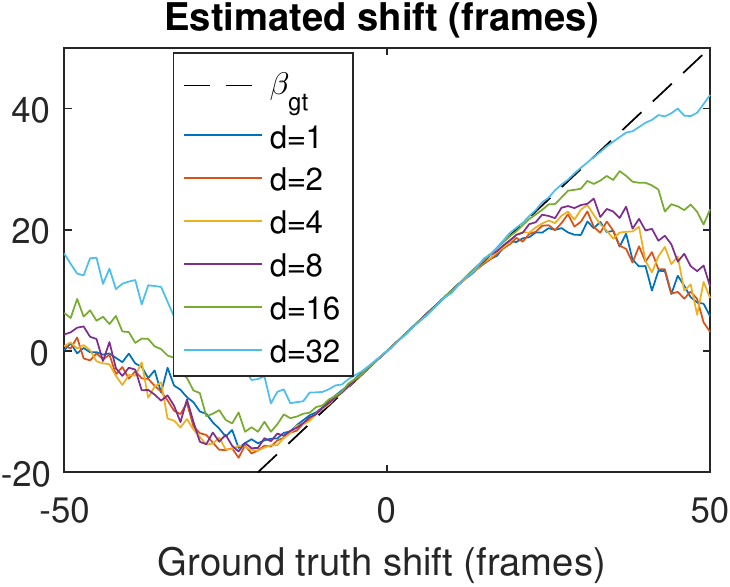}\\
 
\vspace{0.5em}

\includegraphics[width=0.48\columnwidth]{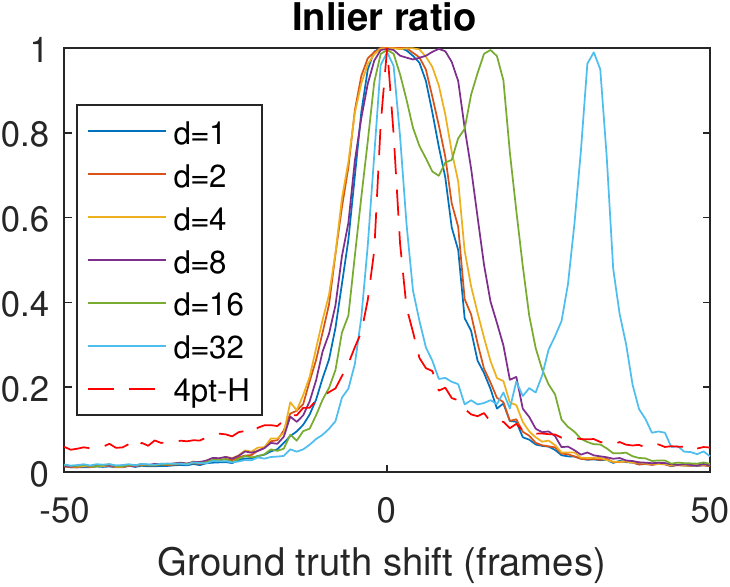}
\includegraphics[width=0.48\columnwidth]{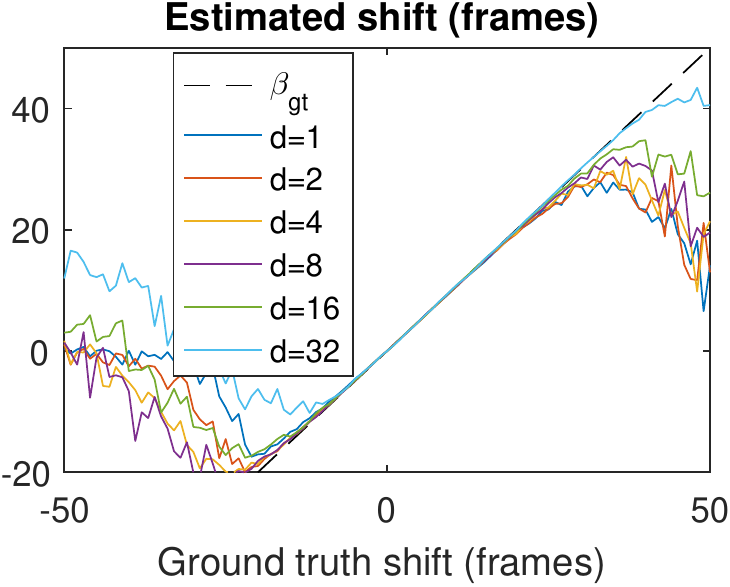}
\caption{Results on randomly generated scene with various time shift $\beta$ between cameras and several different interpolation distances $d$. Temporal distance of one frame equals to approximately 8 pixels distance in 1000x1000px image. Top two figures are results for epipolar matrix and bottom two for homography.}
\label{fig:synth_inliers_beta_F_H}
\end{figure}
There are multiple observations we can make from the results. The main one is that both $\M{F}$ and $\M{H}$ solvers perform well in terms of estimating $\beta_{gt}$, even for the minimal interpolation distance $d=1$. Figure~\ref{fig:synth_inliers_beta_F_H} shows that almost all inliers are correctly classified using $d=1,d=2,d=4$ up to shift of 5 frames forward. Furthermore, even though the inlier ratio begins to decrease with larger shifts, time shift $\beta$ is still correctly estimated, up till frame shifts of 20. Overall, for a given $d$, each algorithm was able to estimate correct $\beta$ at least up to $d$. This is a nice property, suggesting that for larger time shifts we should be able to estimate them simply by increasing $d$. 

For $d=8,d=16,d=32$, the situation is slightly different with respect to inliers. Notice that there are two peaks in the number of inliers, one at $\beta_{gt}=0$ and the other at $\beta_{gt}=d$. This is expected, because at $\beta_{gt}=d$, the interpolating vector $\V{v}$ passes through the sample $\V{s}'_{i+\beta_{gt}}$ which is in temporal correspondence with $\V{s}_i$. When $\beta_{gt} \neq 0$ our solvers are for any $d$ well above the number of inliers provided by standard $\M{F}$ and $\M{H}$ algorithms. 

Another thing to notice is the non-symmetricity of the results. Obviously, when $\beta_{gt}<0$ (backward) and we are interpolating with $d$-th (forward) sample, the peaks in inliers are not present, since we will never hit the sample which is in correspondence. Also, the performance in terms of inliers is reduced when interpolating in the wrong direction, although still above the algorithms not modelling the time shift. Estimation of $\beta$ deteriorates significantly sooner for negative $\beta_{gt}$, at around -10 frames. We will show how to overcome this non-symmetricity by searching over $d$ in both directions using an iterative algorithm.

\section{Iterative algorithm}
\label{sec:iter}
\noindent
As we observed in the synthetic experiments, the performance of the minimal solvers will depend on the distance from the optimum, i.e. the distance between the initial estimate $\beta_0$ and the true time shift $\beta_{gt}$, and on the distance $d$ of the samples used for interpolation. The results from synthetic experiments (Figure~\ref{fig:synth_inliers_beta_F_H}) provide useful hints on how to construct an iterative algorithm to improve the performance and applicability of the minimal solvers. In particular, there are three key observations to consider.

First, the number of inliers obtained from RANSAC seems to be a reasonable function to optimize. Generally it will have two strong local maxima, one at $t_i = t'_i$ and one at $(t_i - t'_i) = d$. At $t_i = t'_i$ the sequences are synchronized and at $(t_i - t'_i) = d$, Fig.~\ref{fig:synth_inliers_beta_F_H}, we obtain the correct $\beta$. Both situations give us synchronized sequences. Second, the $\beta$ computed even far from the optimum, although not precise, provides often a good indicator of the direction towards $t_i = t'_i$. Finally, it can be observed that increasing $d$ improves the estimates when we are far from the optimum. Moreover, as seen from the peaks in Fig.~\ref{fig:synth_inliers_beta_F_H}, selecting larger $d$ yields increasingly better estimates of $\beta$, which are lower or equal than the actual $(t_i - t'_i)$, but never higher. This suggests that we could safely increase $d$ until a better estimate is found.

The observations mentioned above lead us to algorithm~\ref{alg:iter}. The basic principle of the algorithm is the following. In the beginning, assume $i=j$. At each iteration $k$, estimate $\beta$ and $\M{F}$. If this model gives more inliers than previous estimate, change $j$ to the nearest integer to $j+\beta$ and repeat. If the new estimate gives less inliers than the last one, extend the search direction by increasing $d$ by powers of 2 until more inliers are found. If $d^{p_{max}}$ is reached, $p$ is reset to $0$, so interpolation distances keep circling between $d^0$ and  $d^{p_{max}}$. This is essentially a line search over the parameter $d$. Algorithm is stopped when the number of inliers did not increase $p_{max}$ times. This ensures, that at each $t'_j$, all interpolation distances are tested at maximum once. The resulting estimate of $\beta$ is then  $j-i+\beta$, which is the difference in frames the algorithm traveled plus the last estimate of time shift at this point (subframe synchronization).

Estimating of $\beta$ and $\M{F}$ is done using RANSAC and interpolating from both the next and previous $d^{th}$ sample, searching the space of $\beta$ in both directions. Whichever direction returns more inliers is taken as current estimate. By changing the values $p_{min}$ and $p_{max}$ we have the option to adjust the range of search. Having an initial guess about the amount of time shift, e.g.\ not more than 100 frames, but definitely more than 10 frames, we could start the algorithm with values $p_{min}=3$ and $p_{max}=7$ so the search in $d$ would start with $d=8$ and not go further than $d=128$.

\begin{algorithm}
\footnotesize
\caption{Iterative sync}
\label{alg:iterative}
\def\inliers{\textrm{\it inliers}}
\def\skipped{\textrm{\it skipped}}
\newcommand{\sgn}{\mathrm{sgn}}
\begin{algorithmic}
\REQUIRE $\V{s}_0,\ldots,\V{s}_n$,$\V{s}'_0,\ldots,\V{s}'_{n'}$,$k_{max}$,$p_{max}$,$p_{min}$
\ENSURE $\beta$,$\M{T}$
\STATE $\beta_0 \leftarrow 0$,$i=j$,$\skipped \leftarrow 0$, $d \leftarrow 2^{p_{min}}$,$\inliers_0 \leftarrow 0$,$p \leftarrow p_{min}$
\WHILE{$k=1<k_{max}$}
    \STATE $\M{T}_1$,$\beta_1$ and $\inliers_1 \leftarrow$ RANSAC($\V{s}_i,\V{s}_j',d$)  
    \STATE $\M{T}_2$,$\beta_2$ and $\inliers_2 \leftarrow$ RANSAC($\V{s}_i,\V{s}_j',-d$)
    \IF{$\inliers_1 > \inliers_2$}
        \STATE $\inliers_k \leftarrow \inliers_1$, $\beta_k \leftarrow \beta_1$, $\M{T}_k \leftarrow \M{T}_1$
    \ELSE
        \STATE $\inliers_k \leftarrow \inliers_2$, $\beta_k \leftarrow \beta_2$, $\M{T}_k \leftarrow \M{T}_2$
    \ENDIF
    \IF{$\skipped>p_{max}$}
        \RETURN $\M{T}_{k-1}$,$\beta \leftarrow j-i + \beta_{k-1}$
    \ELSIF{$\inliers_k < \inliers_{k-1}$}
        \IF{$p<p_{max}$}
            \STATE $p \leftarrow p+1$
        \ELSE
            \STATE $p \leftarrow 0$
        \ENDIF
        \STATE $d \leftarrow  2^p $
        \STATE $\skipped \leftarrow \skipped+1$
    \ELSE
        \STATE $j \leftarrow j+\round{\beta_k}$
        \STATE $\skipped \leftarrow 0$
        \STATE $k \leftarrow k+1$
    \ENDIF
\ENDWHILE
\end{algorithmic}
\label{alg:iter}
\end{algorithm}

The symbol $\M{T}$ represents a geometric relation, in our case either a fundamental matrix or a homography.

\section{Real data experiments}
\noindent
Our real datasets contain two private datasets and three publicly available multi-camera datasets. We aimed at collecting various types of scenes to cover wide range of applications. The public data were always synchronized and we manually shifted the frame to frame correspondences to simulate the ground truth time shift. We experimented with shift of -50 to 50 frames on each dataset, which produced time shifts ranging from 2s to 5s based on the camera framerate.

\subsection{Datasets}
Dataset Marker was obtained by moving an Aruco marker in front of a two webcams running at 10fps. A digital clock in the scene was processed by OCR in each frame to provide ground truth timestamps. 
Further, we used three public datasets and one private: UvA~\cite{Hoffmann2009}, KITTI~\cite{Kitti}, Hockey and PETS~\cite{Ellis}. The UvA dataset consists of video sequences taken by three static cameras with manual annotations of humans. The KITTI dataset contains stereo video sequences taken from a moving car. In our experiments, we used raw unsynchronized data provided by the authors. The Hockey dataset was synchronized by \cite{Smid2017} and the trajectories are manually curated tracks of \cite{Henriques2012}. The PETS dataset is a standard multi-target tracking dataset. Trajectories were detected by \cite{voc-release5, Felzenszwalb2010, Shi2015} and manually joined.
\subsection{Algorithms}
\begin{figure*}[t]
\setlength{\tabcolsep}{1pt}
\begin{tabular}{ccccccc}
\multicolumn{7}{c}{\includegraphics[height=1.5em]{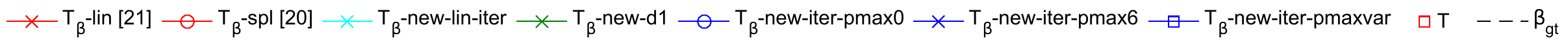} }
\\
& camera 1 trajectories & camera 2 trajectories & \hspace{1em}success rate(\%) & \hspace{1.5em}estimated $\beta$ & \hspace{1.5em}inliers (\%) &
\\
\rotatebox{90}{\hspace{1em}Marker} & 
\includegraphics[width=0.192\textwidth]{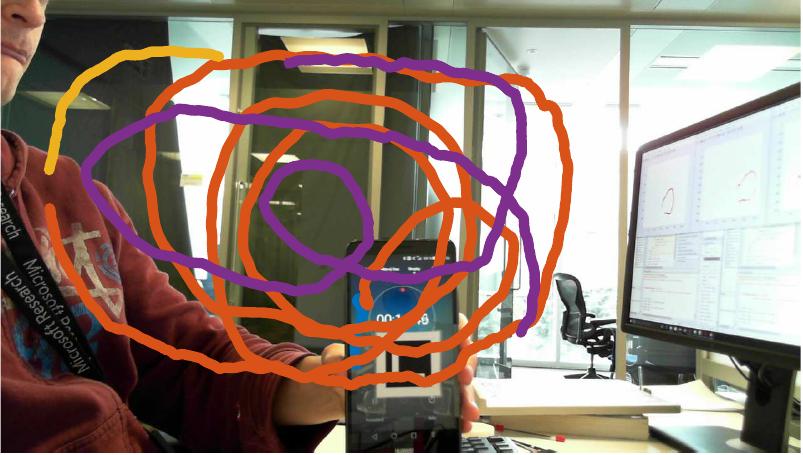} &
\includegraphics[width=0.192\textwidth]{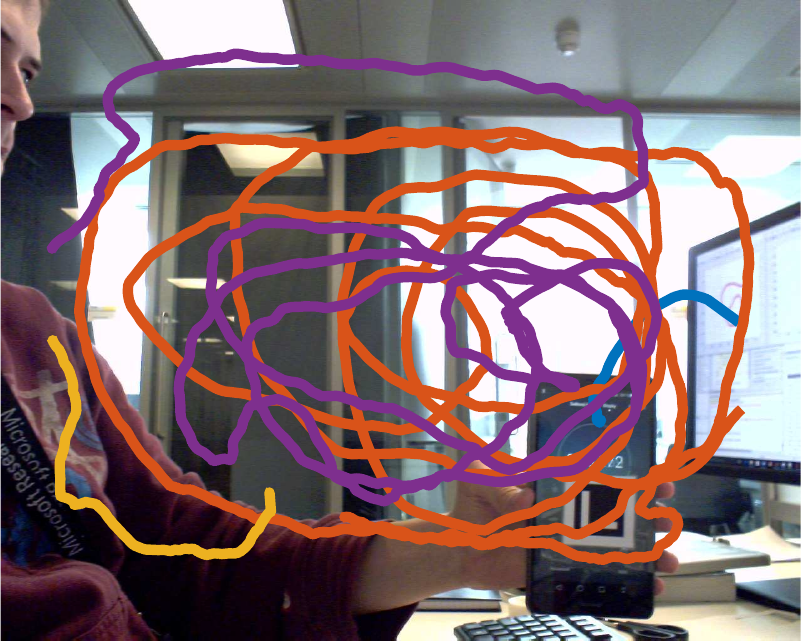} &
\includegraphics[width=0.192\textwidth]{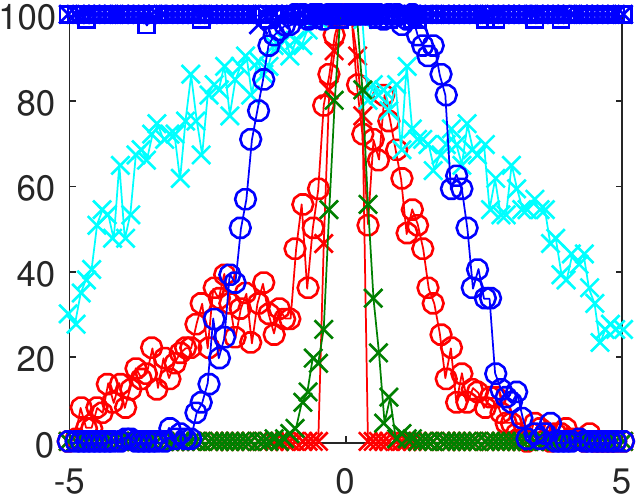}&
\includegraphics[width=0.192\textwidth]{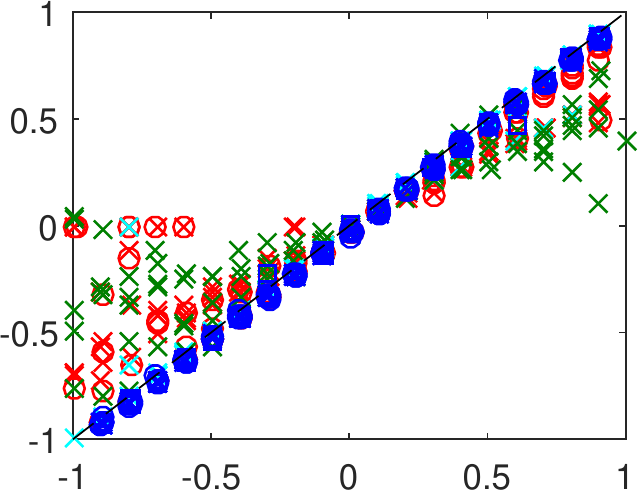}&
\includegraphics[width=0.192\textwidth]{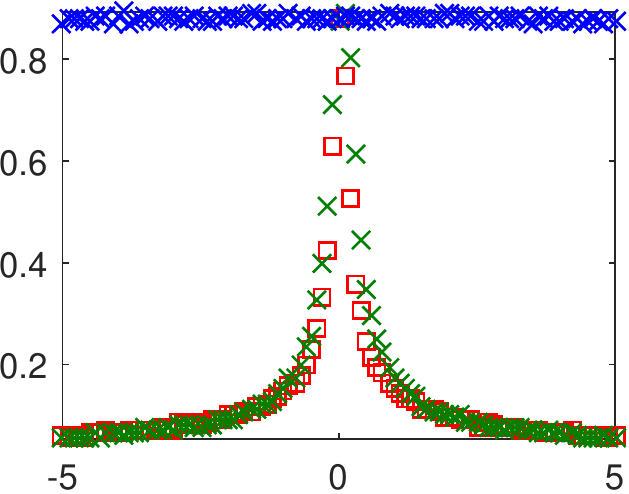}&
\rotatebox{270}{\hspace{-4em}F} 
\\
\rotatebox{90}{\hspace{1em}UvA}& 
\includegraphics[width=0.192\textwidth]{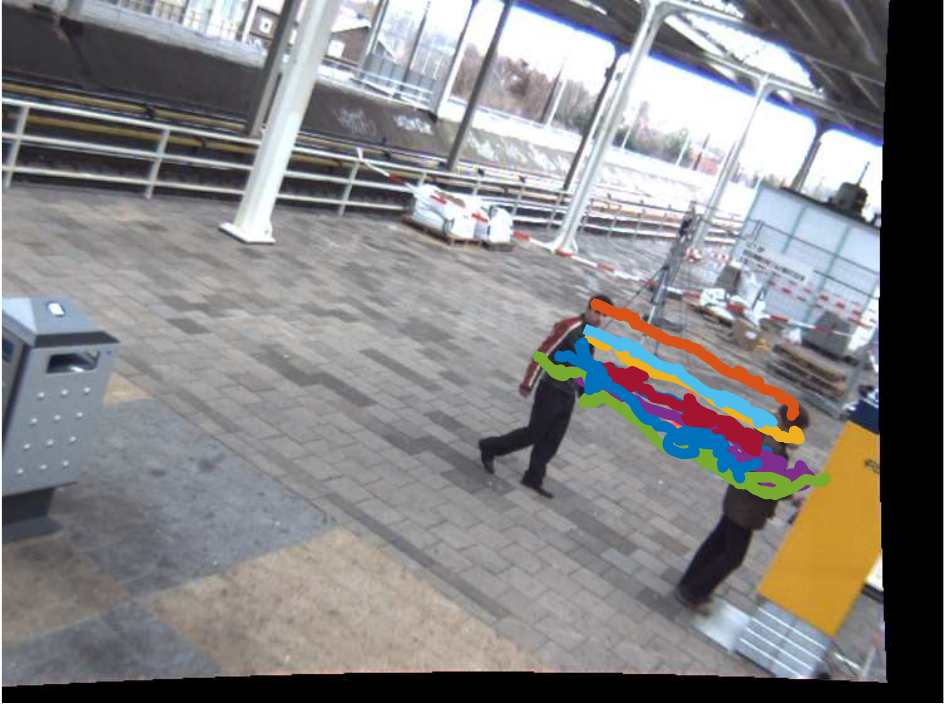}&
\includegraphics[width=0.192\textwidth]{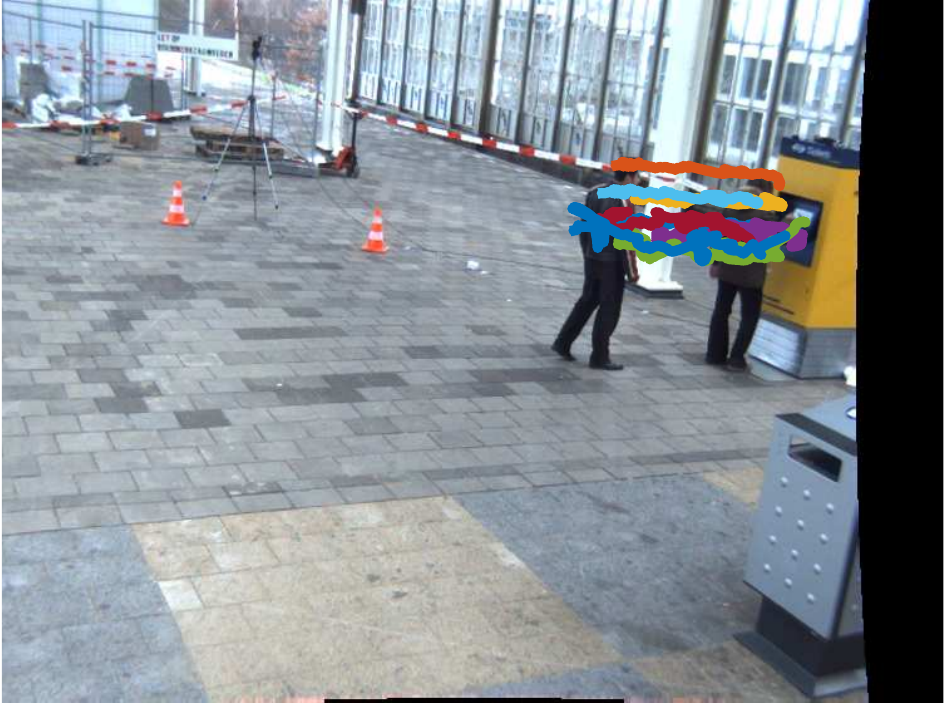}&
\includegraphics[width=0.192\textwidth]{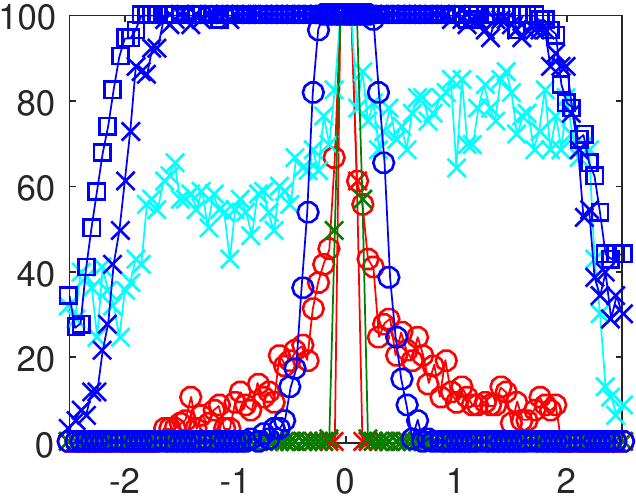}&
\includegraphics[width=0.192\textwidth]{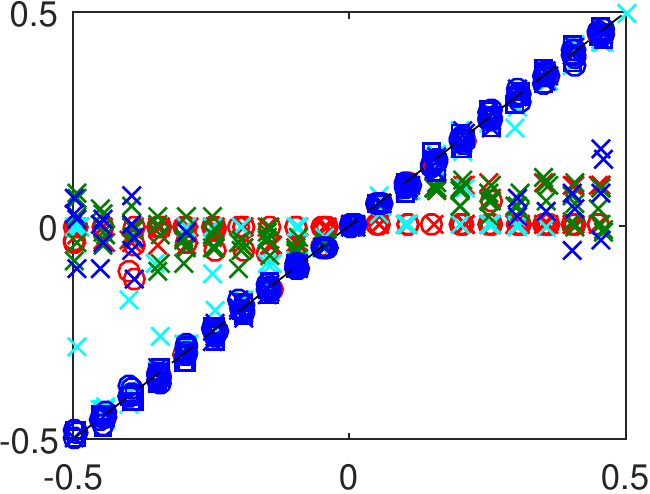}&
\includegraphics[width=0.192\textwidth]{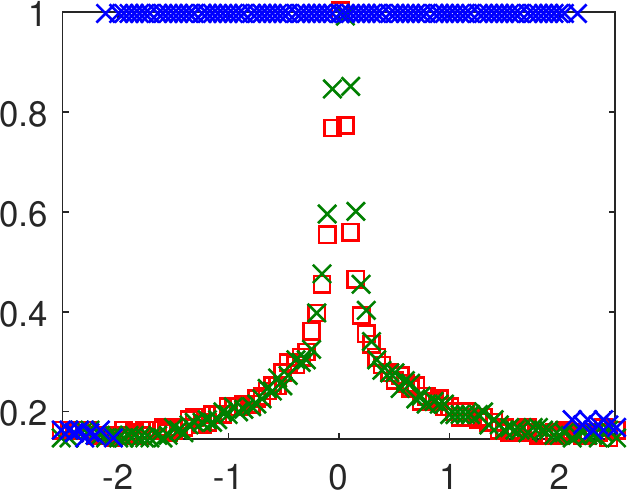}&
\rotatebox{270}{\hspace{-4em}F} 
\\
\rotatebox{90}{\hspace{1em}Kitti}& 
\includegraphics[width=0.192\textwidth]{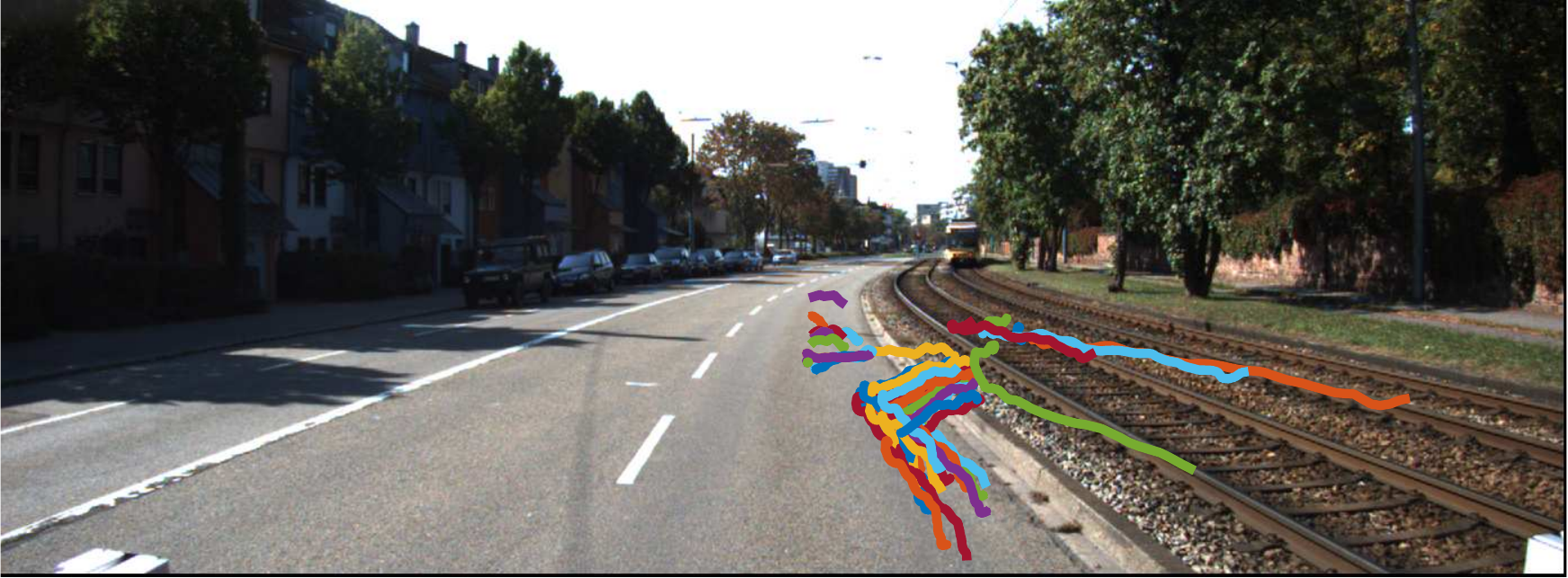}&
\includegraphics[width=0.192\textwidth]{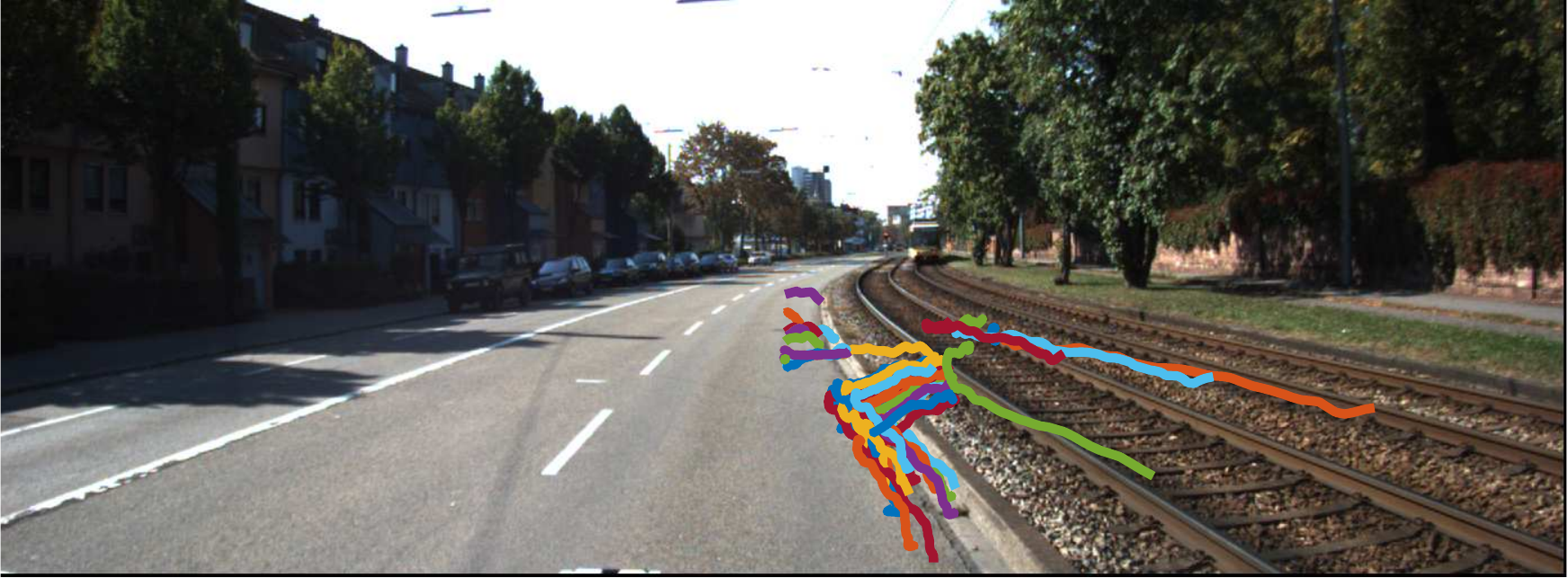}&
\includegraphics[width=0.192\textwidth]{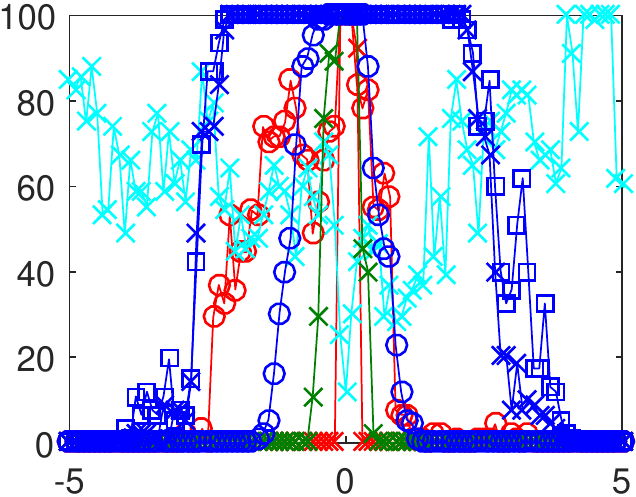}&
\includegraphics[width=0.192\textwidth]{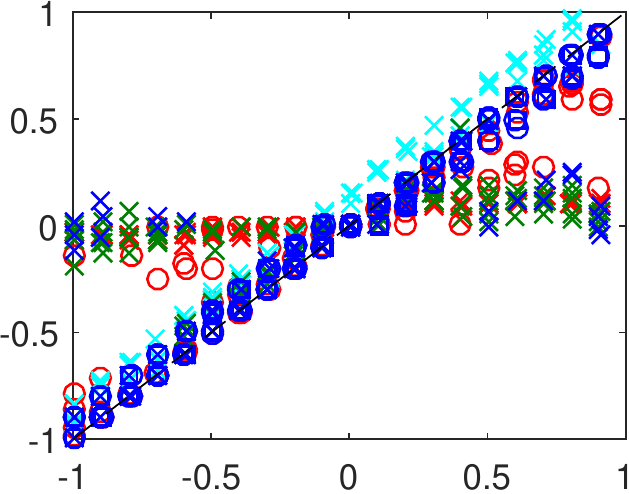}&
\includegraphics[width=0.192\textwidth]{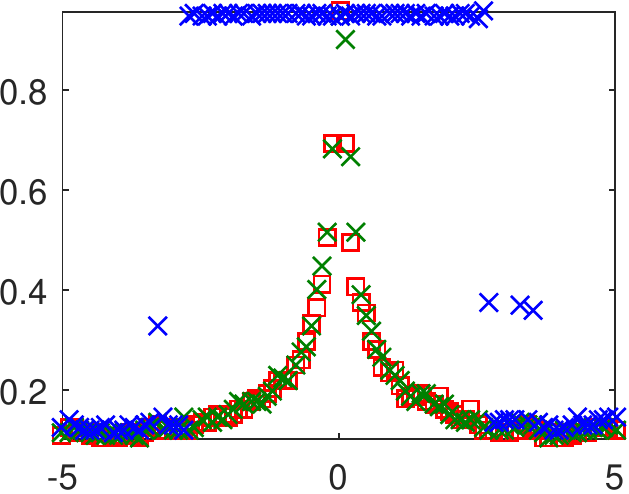}&
\rotatebox{270}{\hspace{-4em}F} 
\\
\rotatebox{90}{\hspace{1em}Hockey}& 
\includegraphics[width=0.192\textwidth]{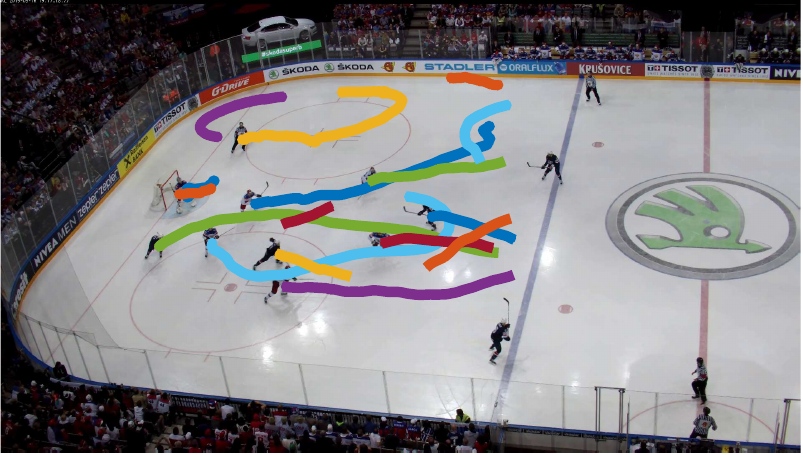}&
\includegraphics[width=0.192\textwidth]{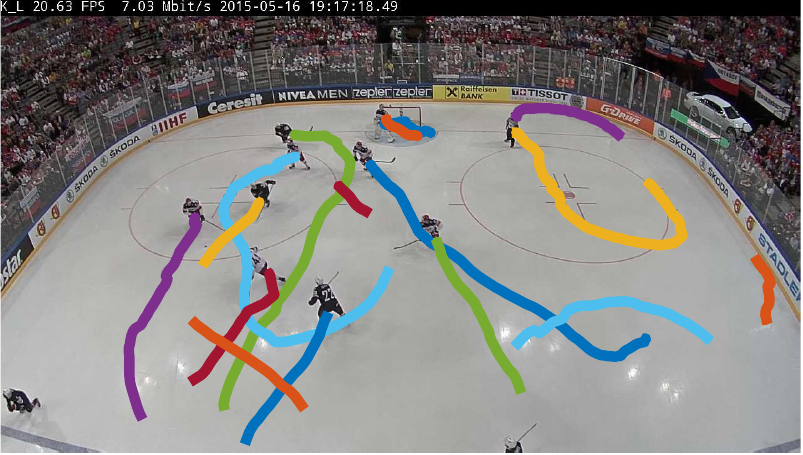}&
\includegraphics[width=0.192\textwidth]{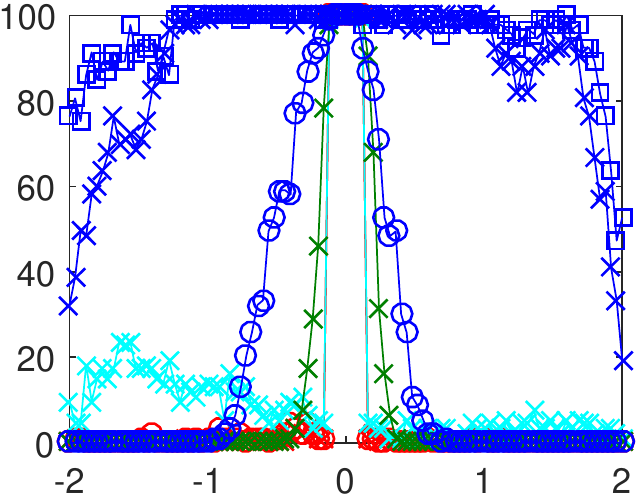}&
\includegraphics[width=0.192\textwidth]{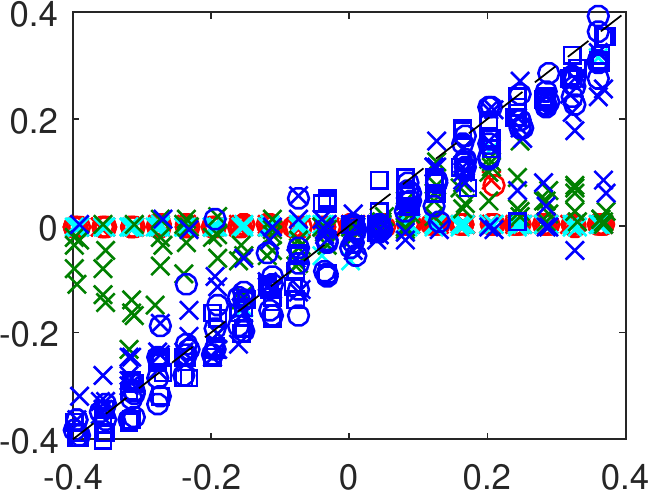}&
\includegraphics[width=0.192\textwidth]{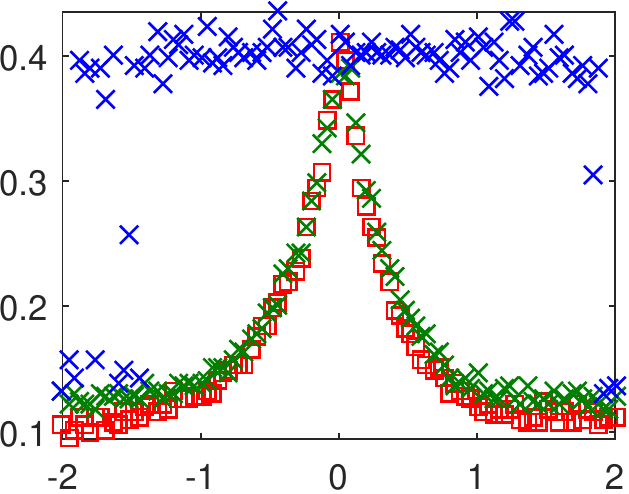}&
\rotatebox{270}{\hspace{-4em}F} 
\\
\rotatebox{90}{\hspace{1em}Hockey}& 
\includegraphics[width=0.192\textwidth]{figs/hokej_img1}&
\includegraphics[width=0.192\textwidth]{figs/hokej_img2}&
\includegraphics[width=0.192\textwidth]{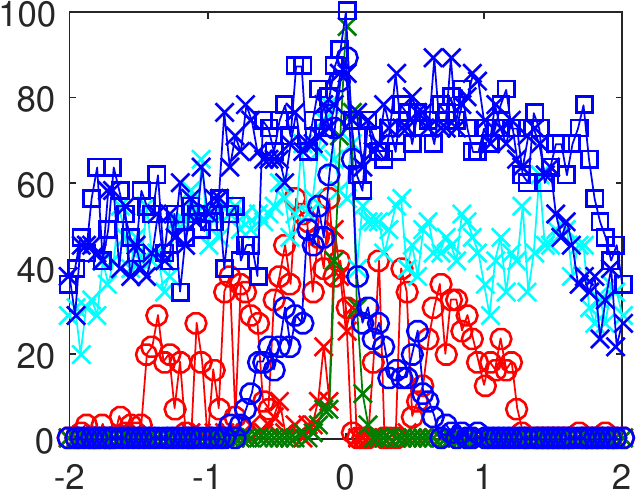}&
\includegraphics[width=0.192\textwidth]{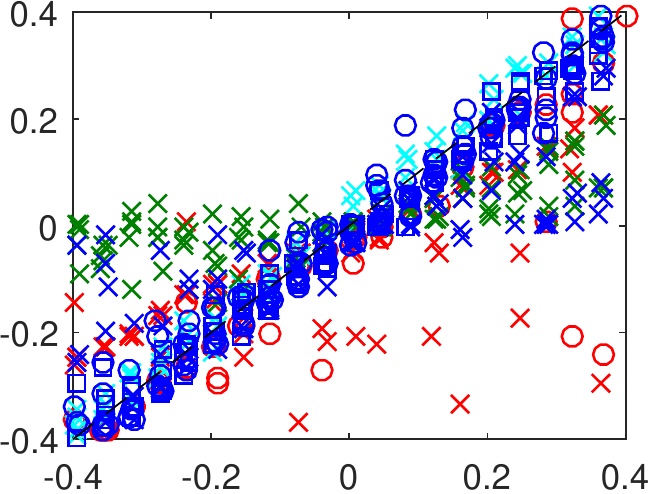}&
\includegraphics[width=0.192\textwidth]{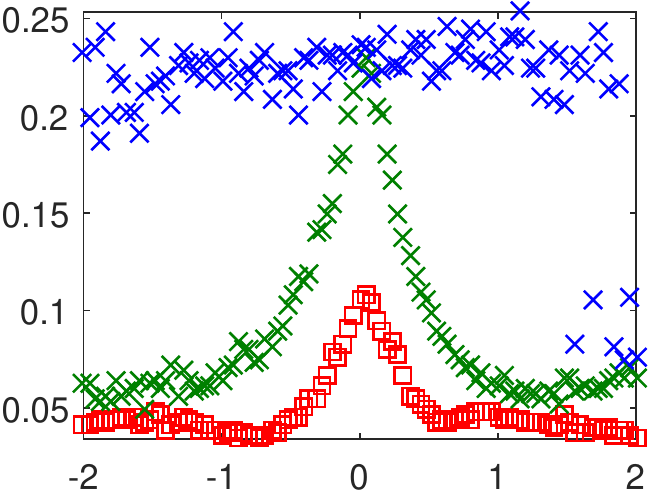}&
\rotatebox{270}{\hspace{-4em}H} 
\\
\rotatebox{90}{\hspace{1em}PETS}& 
\includegraphics[width=0.192\textwidth]{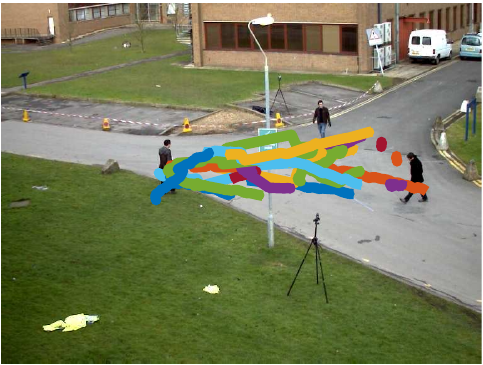}&
\includegraphics[width=0.192\textwidth]{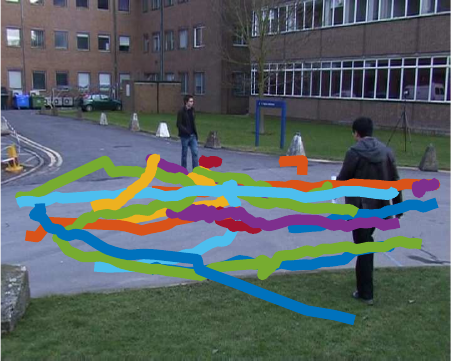}&
\includegraphics[width=0.192\textwidth]{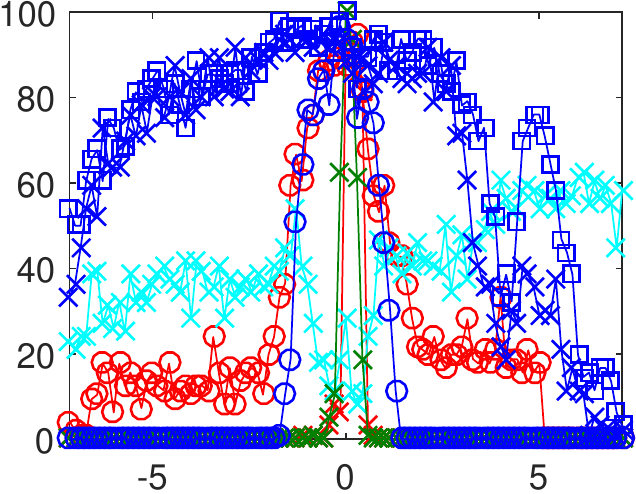}&
\includegraphics[width=0.192\textwidth]{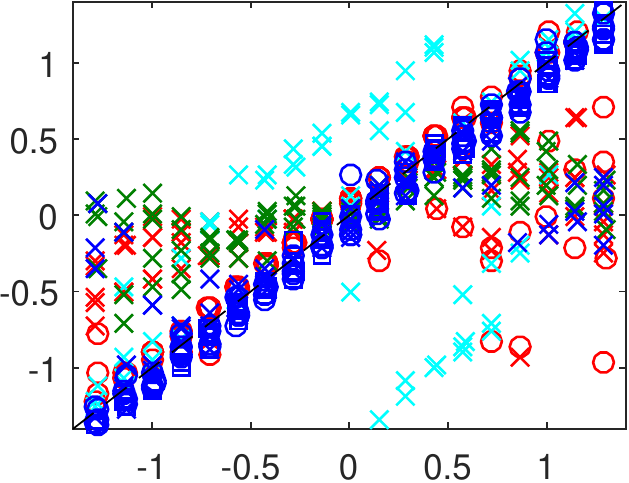}&
\includegraphics[width=0.192\textwidth]{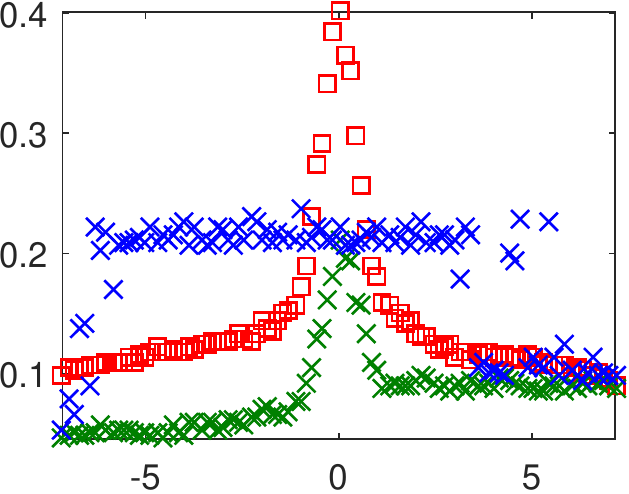}&
\rotatebox{270}{\hspace{-4em}H} 
\\
&  &  & \footnotesize $\beta_{gt}$ (s) & \footnotesize $\beta_{gt}$ (s) & \footnotesize $\beta_{gt}$ (s)
\end{tabular}
\caption{Results on real data. In the two leftmost columns, trajectories used for the computations are depicted in coloured lines over a sample images from the dataset. Third column shows the rates with which different algorithms succeeded to synchronize the sequence to single frame precision for various ground truth time shifts. Fourth column shows a closer look at the individual results for $\beta$ for smaller ground truth time shifts and five runs for each algorithm, each data point corresponds to one run of the algorithm at corresponding $\beta_{gt}$. Letters H and F on the right signalize whether homography or fundamental matrix was computed. }
\label{fig:real_data}
\end{figure*}

We compared seven different approaches to simultaneously solving two-camera geometry and time shift. Depending on the data, either fundamental matrix or homography was estimated. We denote both geometric relations by $T$, where $T$ means $\M{H}$ or $\M{F}$ was estimated using standard 4 or 7 point algorithms~\cite{hartley2003} and $T_\beta$ means that $\M{H}$ or $\M{F}$ was estimated together with $\beta$. 
The rightmost column of figure~\ref{fig:real_data} shows which model, i.e.\ homography or fundamental matrix, was estimated on a particular data set.

The closest alternatives to our approach are the least-squares based algorithms presented in~\cite{Noguchi2007} and~\cite{Nischt2009}. Both optimize $\M{F}$ or $\M{H}$ and $\beta$ starting from an initial estimate of $\beta=0$ and $T$. Method~\cite{Noguchi2007} uses linear interpolation from the next sample, whereas method~\cite{Nischt2009} uses spline interpolation of the image trajectory and we will refer to these methods as $T_\beta$-lin and $T_\beta$-spl respectively. In our implementation of those methods, we used Matlab's lsqnonlin function with Levenberg-Marquardt algorithm, all stopping criteria set to epsilon and maximum number of 100 iterations. 

We tested the solvers presented in section~\ref{sec:solver} with $d=1$ as algorithm $T_\beta$-new-d1. The proposed iterative algorithm~\ref{alg:iterative} that uses the solvers was tested using several different settings. The user can control the algorithm using parameters $p_{max}$ and $p_{min}$, which determine the distances $d$ that will be used for interpolation. 
As we observed in section~\ref{sec:exp_synthetic}, there is a good chance of computing a correct $\beta$ if $d>\beta_{gt}$. First, we ran the algorithm with $p_{min}=0$ and $p_{max}=6$, which gives maximum $d=64$ as algorithm $T_\beta$-new-iter-pmax6. This version of the algorithm is guaranteed to try $d={1,2,4,8,16,32,64}$ at each $\beta_k$ before it stops or it finds more inliers. This covers the time shifts we tested, but can lead to unnecessary iterations for smaller shifts. Therefore, we also tested $p_{max}=0$ as $T_\beta$-new-iter-pmax0 which only tried $d=1$ at each iteration to see the capabilities of the most efficient version of the algorithm. 

The last version of our algorithm, $T_\beta$-new-iter-pmaxvar, adapted both $p_{max}$ and $p_{min}$ to $\beta_{gt}$ such that $2^{p_{min}}\leq \beta_{gt}<2^{p_{max}}$. This represents a case when user has a rough estimate about the expected time shift and sets the algorithm accordingly. We remind that setting  $p_{min}$ only affects the initial interpolating distance, after reaching $d=2^{p_{max}}$ the algorithm starts again with $d=2^0$.

Finally, algorithm $T$-lin~\cite{Noguchi2007} also takes the next samples for interpolation, making it comparable to our $T_\beta$-new-d1. We used $T$-lin in the same iterative scheme as $T_\beta$-new-iter-pmax6 and tested it as $T$-new-lin-iter, where instead of using the number of inliers as a criteria for accepting a step, we used the value of the residual.

\begin{table}[]
    \centering
    \setlength{\tabcolsep}{1pt}
    \begin{tabular}{c|c|c|c|c|c}
    $\beta_{gt}$ & 0-10  & 10-20 & 20-30 & 30-40 & 40-50 \\ \hline
    $T_\beta$-new-iter-pmax0  & 4.7 & 4.3 & 3.5 & 4.1 & 3.8\\
    $T_\beta$-new-iter-pmax6  & 23 & 22 & 21.2 & 21.6 & 21.2\\
    $T_\beta$-new-iter-pmaxvar  & 18 & 19 & 17.5 & 16.7 & 16.5 \\  \hline
    \end{tabular}
    \vspace{0.1em}
    \caption{Average number of RANSACs executed before termination. Evaluated on Marker dataset.}
    \label{tab:ransacs}
\end{table}

\subsection{Results and discussion}
The results on real datasets demonstrate a wide practical usefulness of the proposed methods. For most datasets, $T_\beta$-new-d1 itself performed at least as good as the least squares algorithms $T_\beta$-lin and $T_\beta$-spl. A single RANSAC was enough to synchronize time shifts of 2-5 frames across all datasets. The iterative algorithm $T_\beta$-new-iter-pmax6 built upon our solvers performed the absolute best across all datasets, converging successfully from as far as 5s time difference on Marker and Hockey datasets, 2s difference on UvA dataset and 2.5s on Kitti dataset as seen in the success rate column of figure~\ref{fig:real_data}. 

On the Kitti dataset,  $T_\beta$-new-iter-pmax6 was outperformed by the  $T_\beta$-new-lin-iter, which uses the iterative algorithm proposed by us, but with a solution from~\cite{Noguchi2007} inside.  $T_\beta$-new-lin-iter was able to estimate time differences larger than 2.5s but only in roughly half of the cases, where  $T_\beta$-new-iter-pmax6 was 100\% successful up to 2.5s when it sharply fell off. We account this to the high non-linearity of the 2D velocity of the image points, where as the objects got closer to the car, they moved faster. The tracks of length 25 frames and more were very sparse here and the longer they were the more non-linear in the velocity. 

On the contrary, the hockey dataset posed a big challenge for the least squares algorithms, which struggled even with the smallest time offsets. We attribute this to the poor estimate of $\M{F}$ by the seven point algorithm which causes the LM algorithm to get stuck in local minima. We also tested the homography version of all algorithms on this dataset, since the trajectories are approximately planar, which resulted in the least squares algorithms performing slightly better whereas the algorithms with minimal solvers performed slightly worse.

PETS dataset was probably the most challenging, because of the low framerate (7FPS), coarse detections and abrupt change of motion. Still, our methods managed to synchronize the sequences in majority of cases.

Table~\ref{tab:ransacs} shows the average number of RANSACs used before termination of different variants of iterative algorithm~\ref{alg:iter} for the dataset Marker. We can see that using 8pt-iter-pmax0 greatly reduces the computations needed, still allowing this method to reliably estimate time shifts of 0.5s-2s depending on the scene, rendering it useful if we are certain that the sequences are off by only a few tens of frames. Knowing the time shift approximately and setting $p_{max}$ and $p_{min}$ can also reduce the computations as shown by 8pt-iter-pmaxvar, which provided identical performance to 8pt-iter-pmax6, sometimes even outperforming it.

\section{Conclusion}
\noindent We have  presented solvers for simultaneously estimating epipolar geometry or homography and time shift between image sequences from unsynchronized cameras. These are the first minimal solutions to these problems, making them suitable for robust estimation using RANSAC. Our methods need only trajectories of moving points in images, which are easily provided by state-of-the-art methods, e.g.\ SIFT matching, human pose detectors, or pedestrian trackers. We were able to synchronize wide range of real world datasets shifted by several frames using a single RANSAC with our solvers. For larger time shifts, we proposed an iterative algorithm using these solvers in succession.  The iterative algorithm proved to be reliable enough for synchronizing real world camera setups ranging from autonomous cars, surveillance videos, and sport game recordings, which were de-synchronized by several seconds. 
\section{Appendix}
\noindent The appendix contains additional experimental results that provide details that are out of the scope of the main paper.

\subsection{Subframe synchronization}
\begin{figure}
    \centering
    \includegraphics[width=\columnwidth]{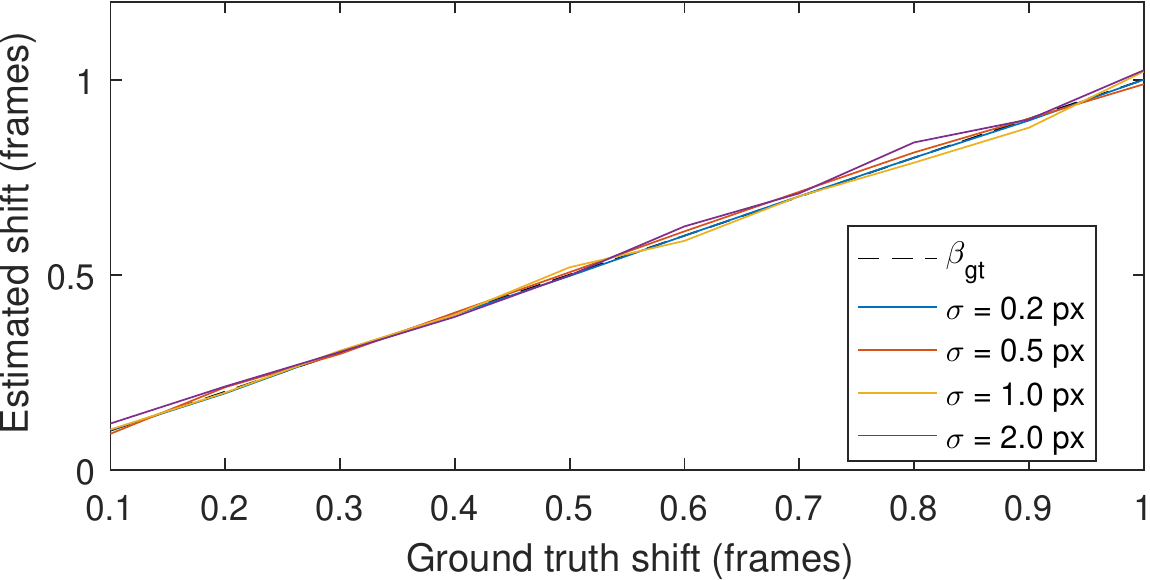}
    \caption{Subframe time shift estimation using the fundamental matrix solver. The solver was tested with different levels of image noise.}
    \label{fig:synth_subframe}
\end{figure}

\begin{figure}
    \vspace{-0.3cm}
    \centering
    \includegraphics[width=0.49\columnwidth]{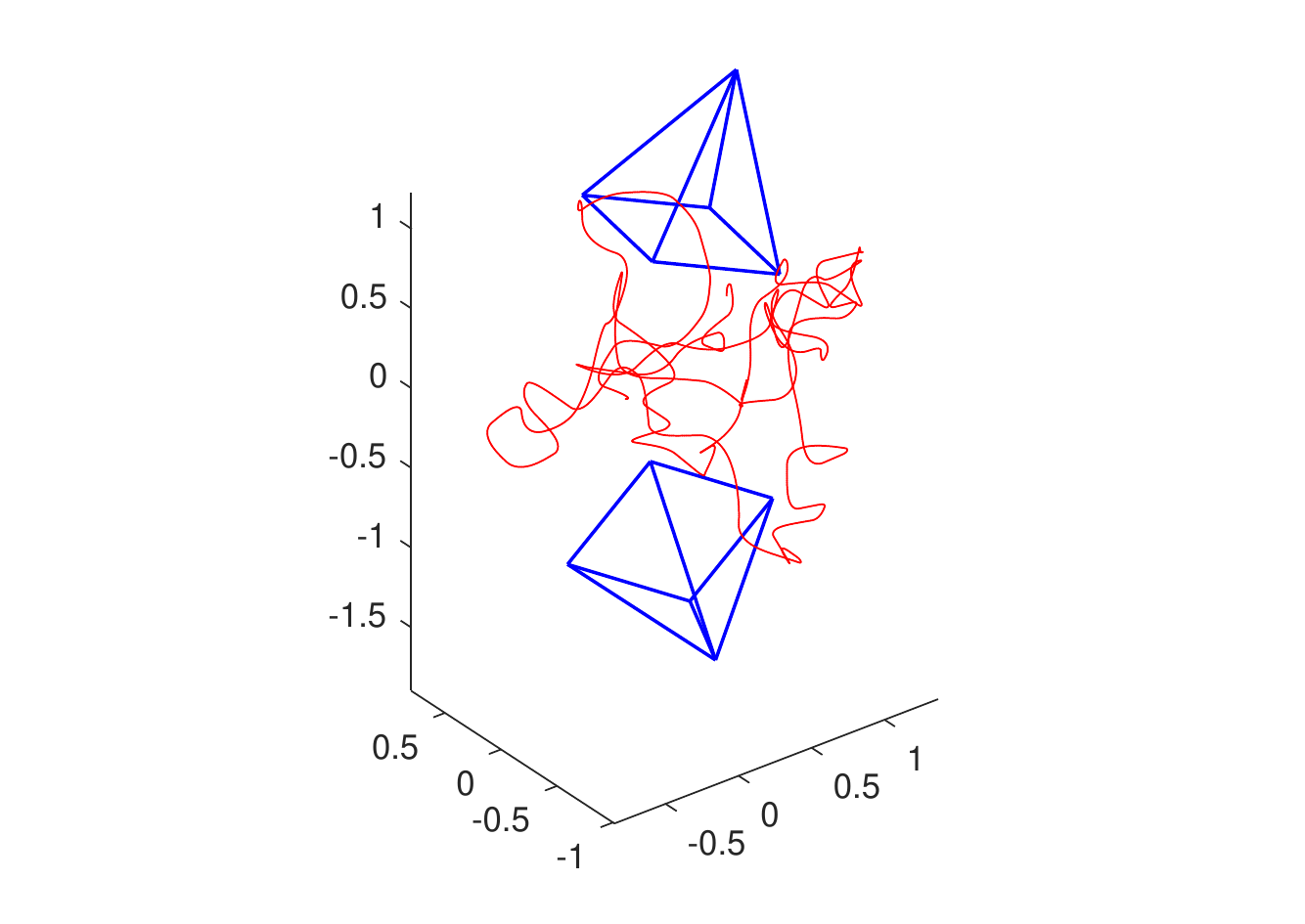}
    \includegraphics[width=0.49\columnwidth]{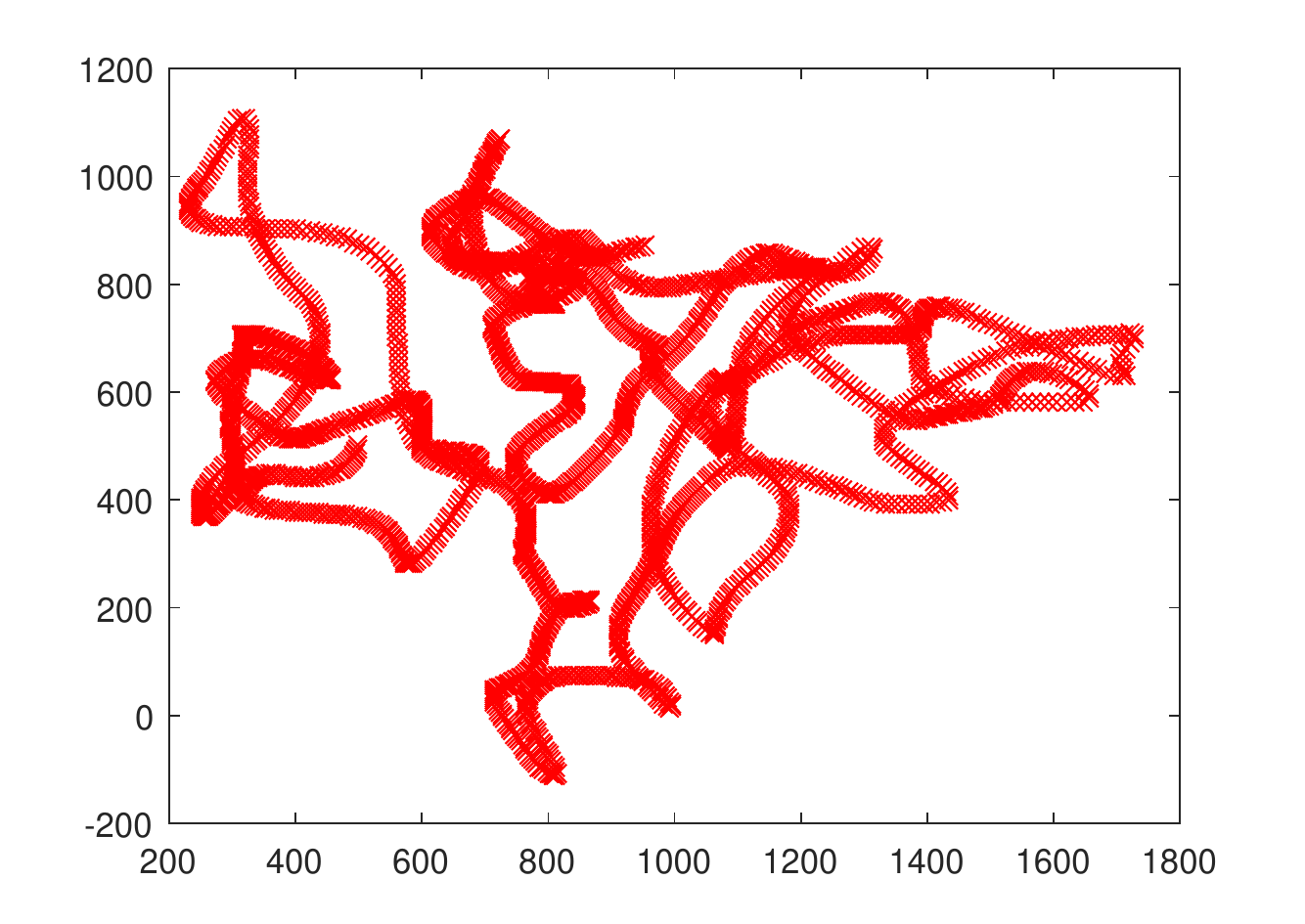}
    \caption{An example of the randomly generated scene for the synthetic experiments. On the left is the 3D trajectory with cameras and on the right is an image projected into one of the cameras.}
    \label{fig:synth_scene}
\end{figure}

\noindent One issue that was not directly elaborated upon in the main paper is the ability of the solvers to synchronize sub-frame time shifts, \ie, shifts where \ $\beta_{gt}$ is not an integer. In the real datasets, images were either hardware synchronized, \ie, $\beta_{gt}=0$, or we did not have precise enough ground truth information about the subframe time shift. Therefore, we tested the subframe synchronization on the synthetic data only. The results in Figure~\ref{fig:synth_subframe} show  that the subframe synchronization is very precise for various levels of noise. Figure~\ref{fig:synth_scene} shows an example of a randomly generated scene for the synthetic experiments.

\subsection{Iterative algorithm visualization}
\begin{figure}
    \centering
    \includegraphics[width=0.95\columnwidth]{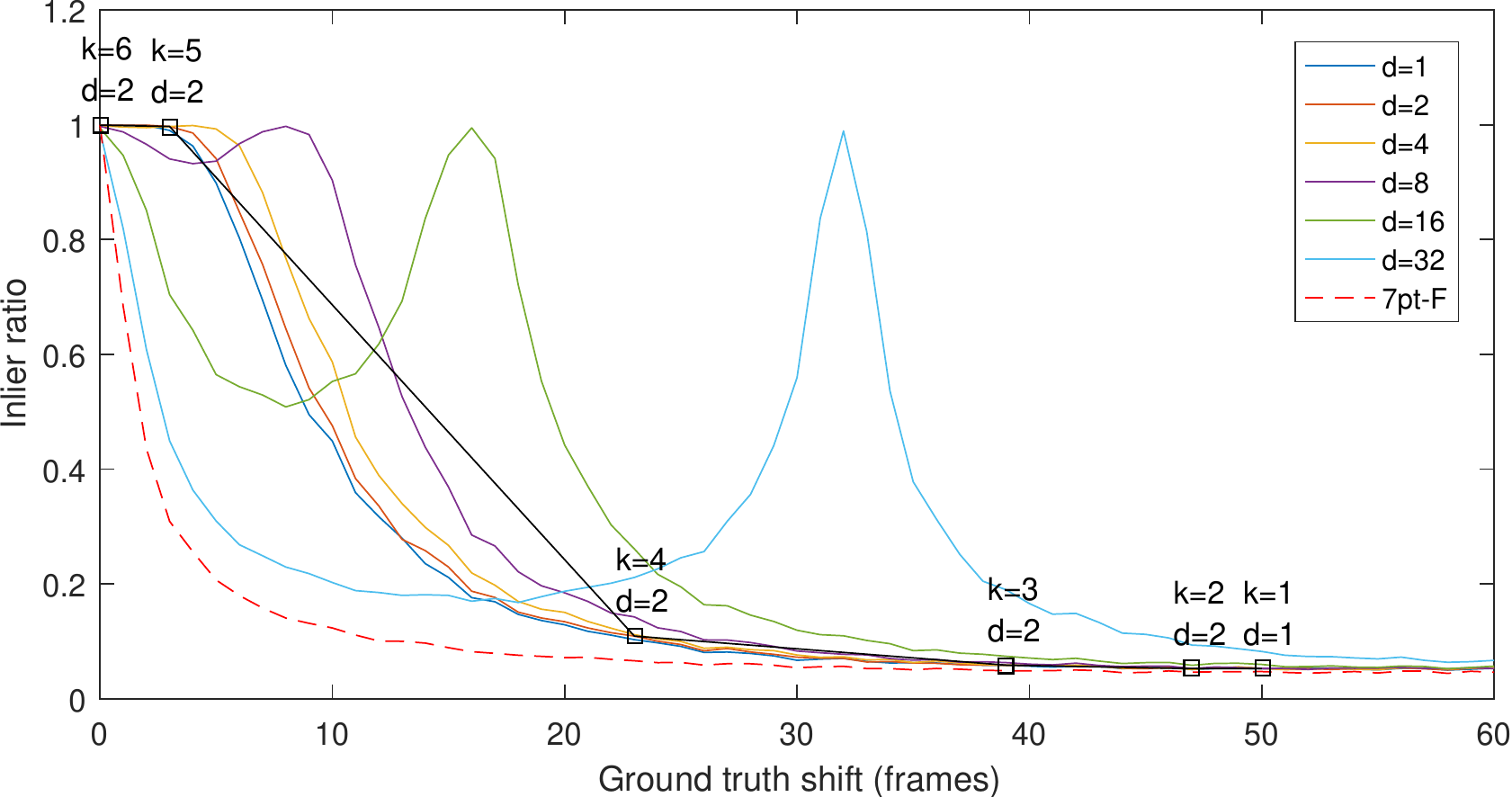}\\
    \includegraphics[width=0.95\columnwidth]{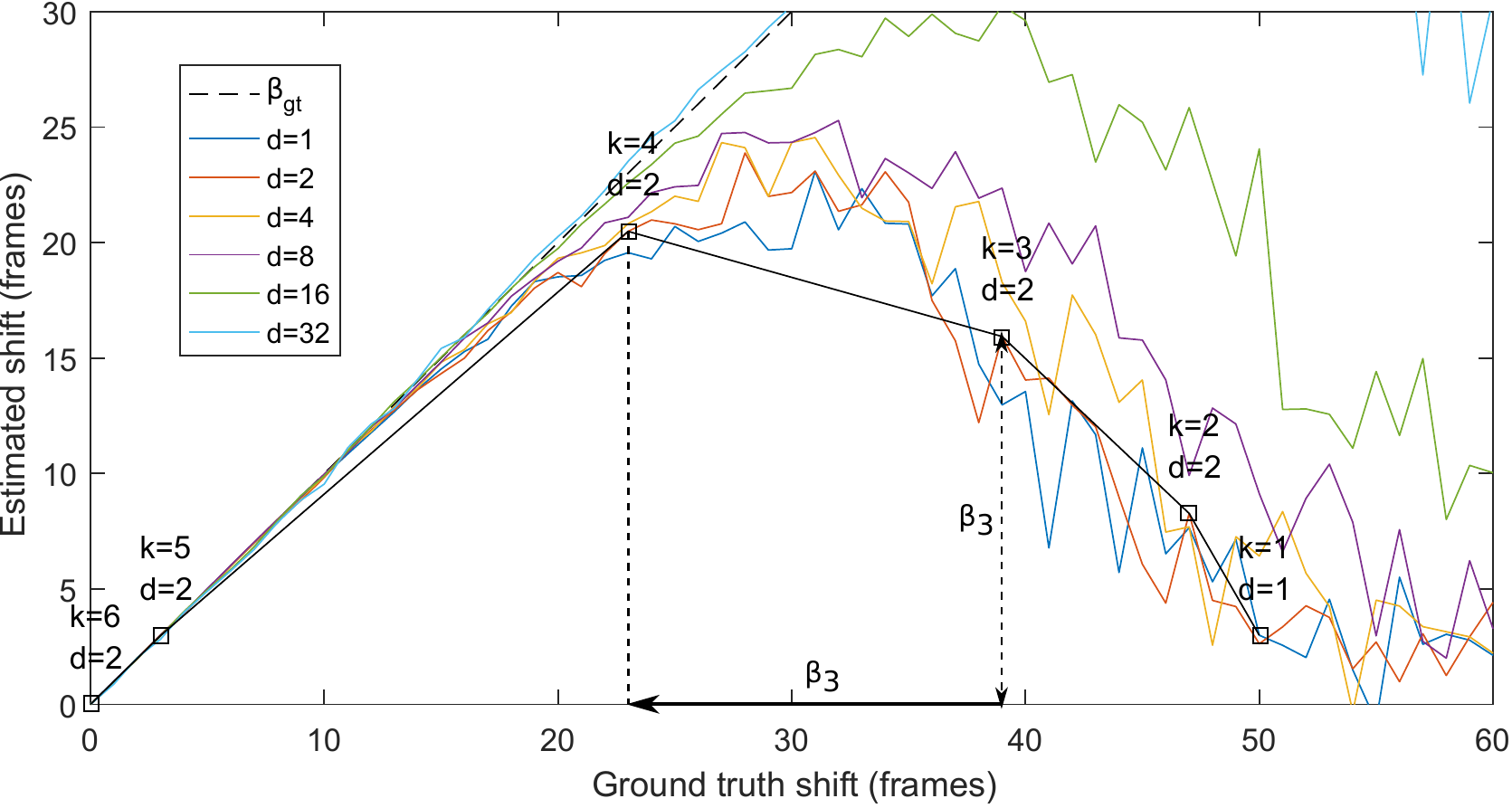}
    \caption{An example of one run of the iterative algorithm. $k$ is the iteration number and $d$ is the interpolation distance used. Beginning with time shift of 50 frames, the algorithm would converge in 6 iterations.}
    \label{fig:iter_run}
\end{figure}

\noindent In Figure~\ref{fig:iter_run}, we provide a visualization of one run of the iterative algorithm with $p_{max}=5$. Each iteration is marked by a black square and denoted by the number of the iteration $k$ and the distance $d$ used for interpolation in the given iteration. The algorithm greedily searches for a larger number of inliers (the top figure) and uses the estimated $\beta_k$ to change the correspondences, which results in change of the current ground truth shift (bottom figure). This particular run converged in 6 iterations, even though the initial time shift (50) was larger than the maximum interpolation distance $d=32$. Moreover, the algorithm only used interpolation distances $d=1,2$. These distances were enough to provide a good enough estimate of the time shift that lead to an increased number of inliers.
\subsection{Accuracy of the estimated geometry}
\subsubsection{Synthetic data}
\begin{figure*}[t]
    \centering
    \includegraphics[width=0.95\columnwidth]{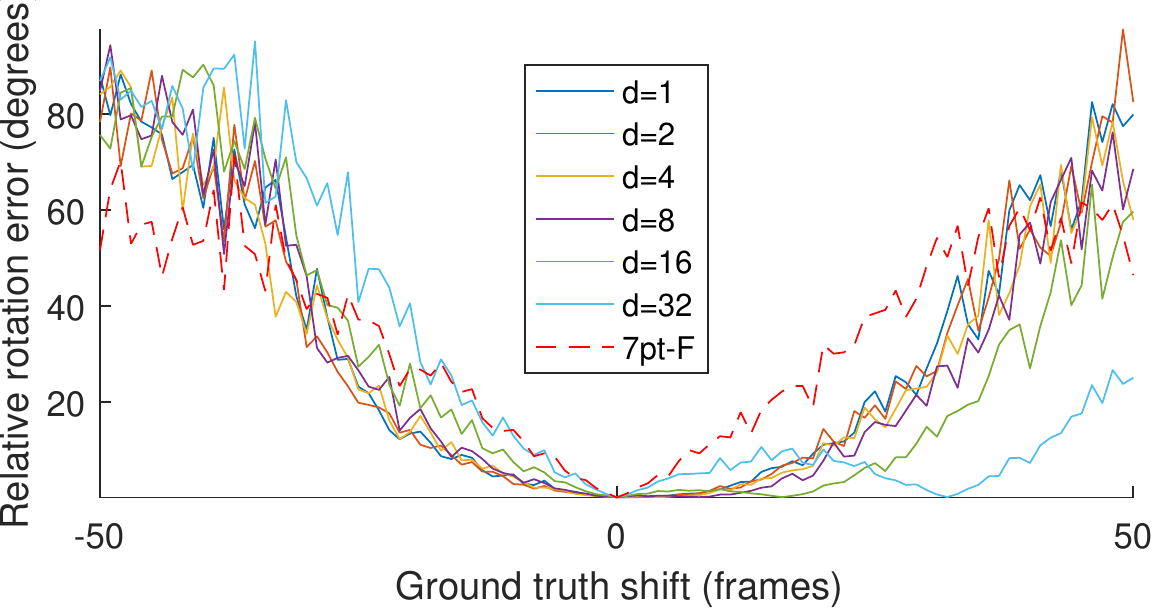}
    \includegraphics[width=0.95\columnwidth]{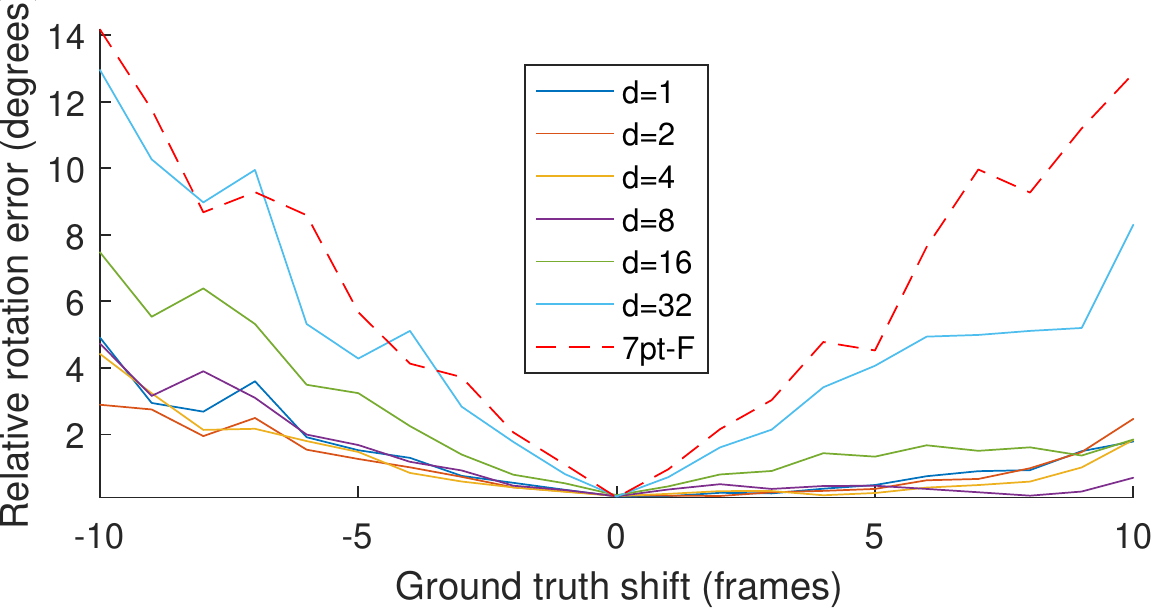}
    \includegraphics[width=0.95\columnwidth]{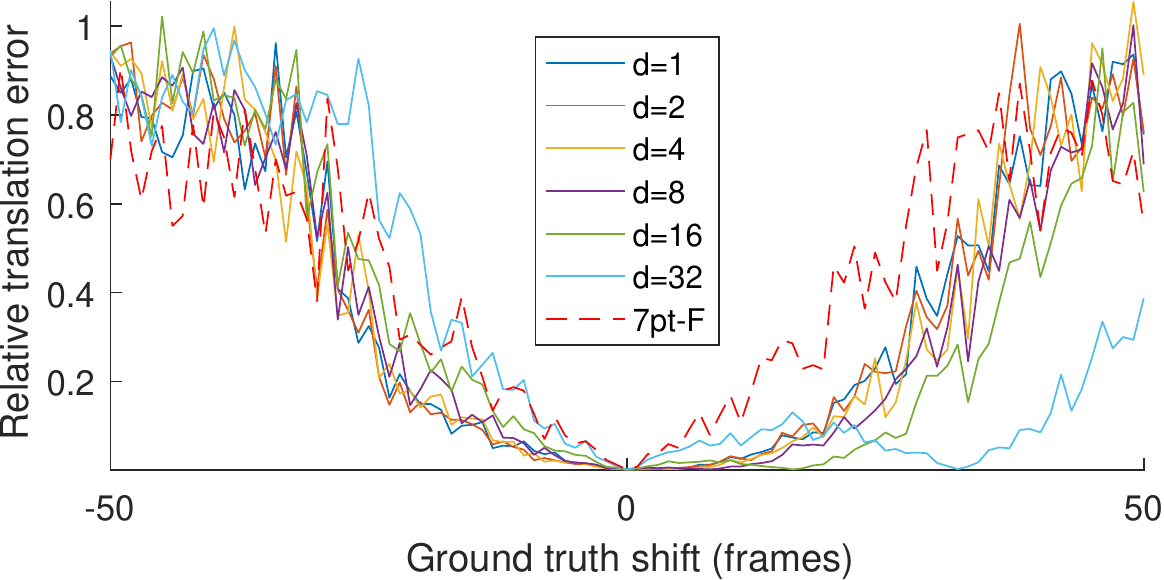}
    \includegraphics[width=0.95\columnwidth]{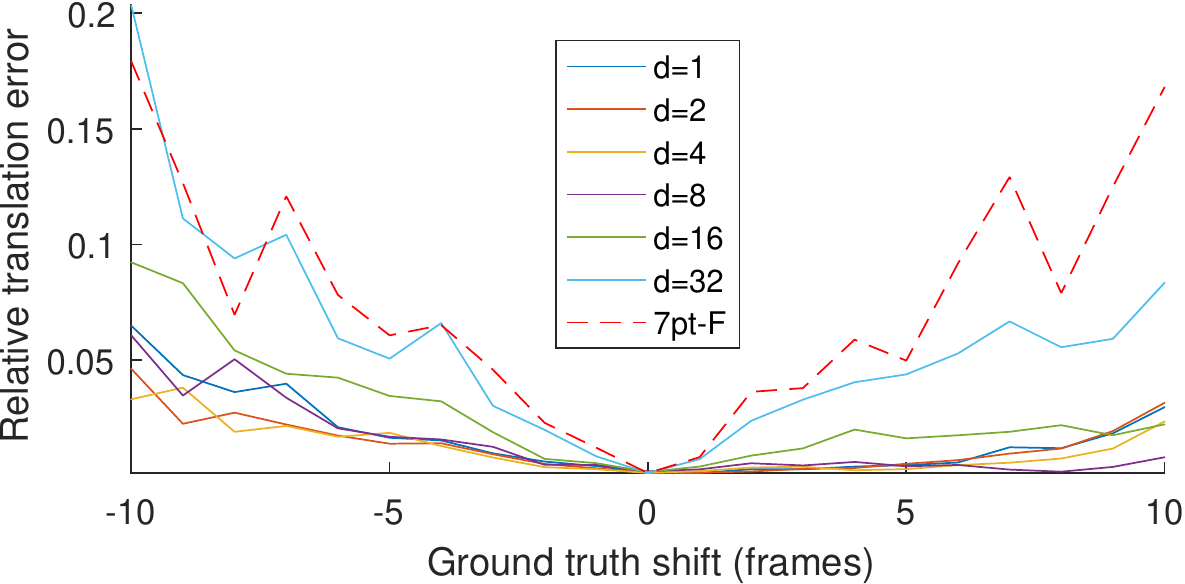}
    \caption{Error in the relative rotation and translation between the two cameras from synthetic data extracted from the computed fundamental matrix. Our solvers provide significantly better rotation and translation estimates than the classic 7-point algorithm. Note that in our iterative algorithm, we only use the right hand side of the results in above graphs, because both $d$ and $-d$ are used at each iteration.}
    \label{fig:synth_Rt_F}
\end{figure*}

\noindent On the same data as used in Section~5.1 of the main paper, we evaluated the estimated relative rotations $\M{R}$ and translations $\V{t}$. The results in Figure~\ref{fig:synth_Rt_F} show that we are able to estimate $\M{R}$ and $\V{t}$ significantly better than the classical 7 point algorithm. The utility of our solver is especially apparent from the zoomed in figures with smaller time shifts. The error in $\M{R}$ and $\V{t}$ is almost zero up to 5 frames shift, for shorter interpolation distances $d=1,2,4,8$. In contrast, such shift causes a significant drop in performance of the classical 7-point algorithm, resulting in errors up to 5 degrees in orientation and relative error of 5\% in the translation vector. 

Even for the long interpolation distances $d=16,32$---although not as good as for $d=1,2,4,8$---the performance is still better than that of the classical 7-point algorithm. The performance of $d=16$ and $d=32$ improves with increasing ground truth time shift and peaks, as expected, on time shifts 16 and 32, respectively. Note that in our iterative algorithm, we only use the right hand side of the results in the above graphs, because both $d$ and $-d$ are used at each iteration.

\subsubsection{Real data}
\begin{figure*}[t]
    \centering
    \includegraphics[width=0.8\paperwidth]{figs/legend1}
    \includegraphics[width=\columnwidth]{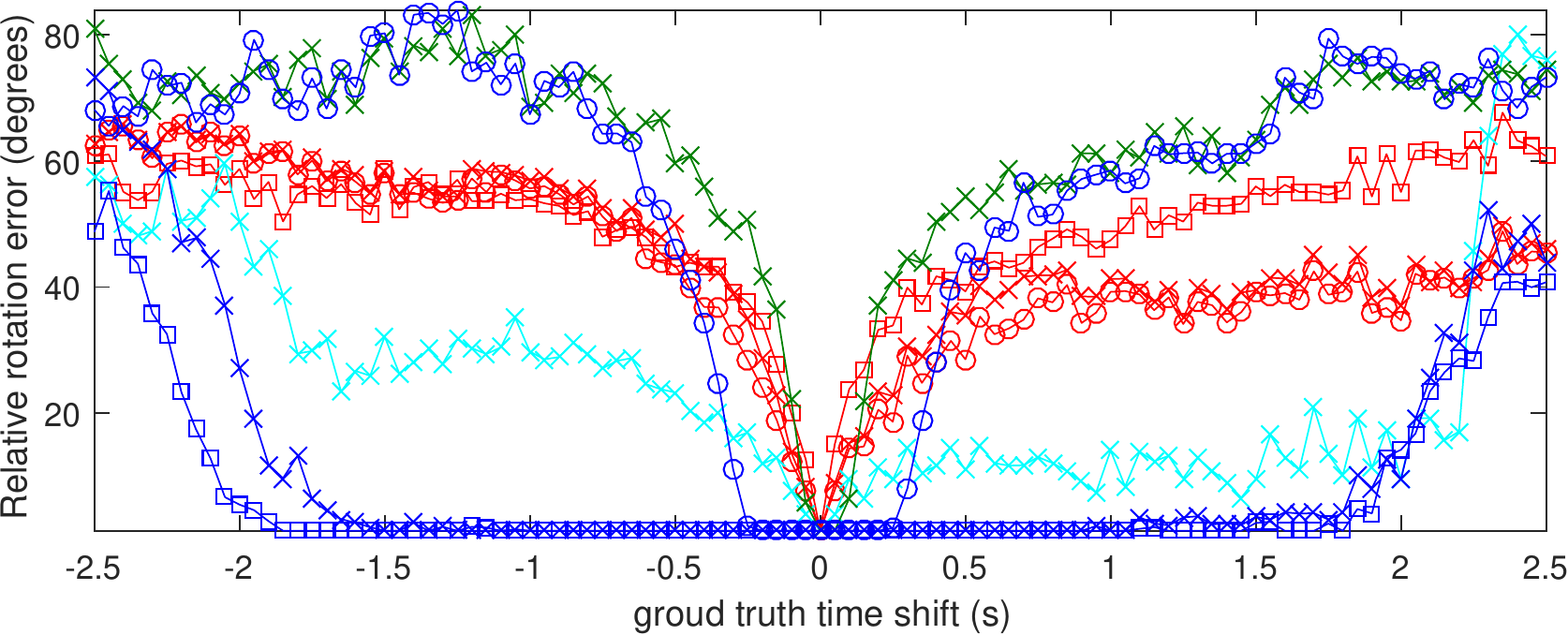}
    \includegraphics[width=\columnwidth]{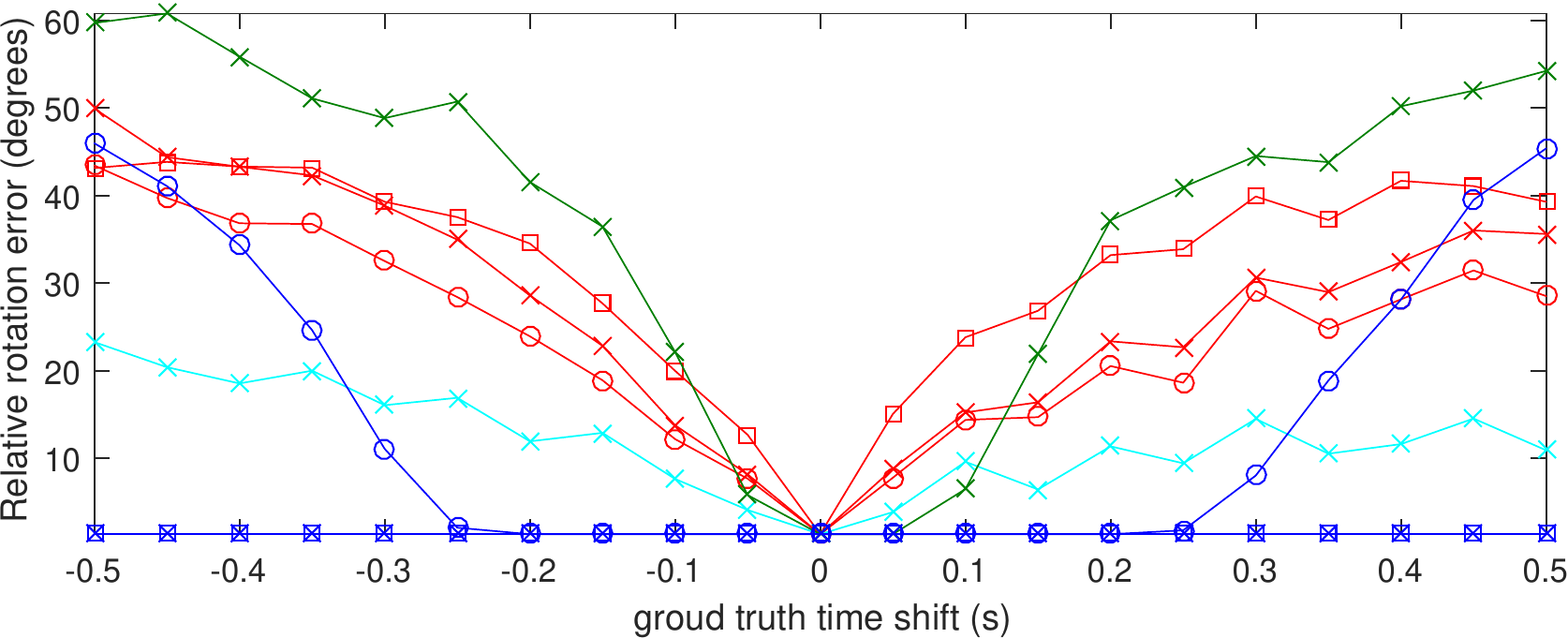}
    \includegraphics[width=\columnwidth]{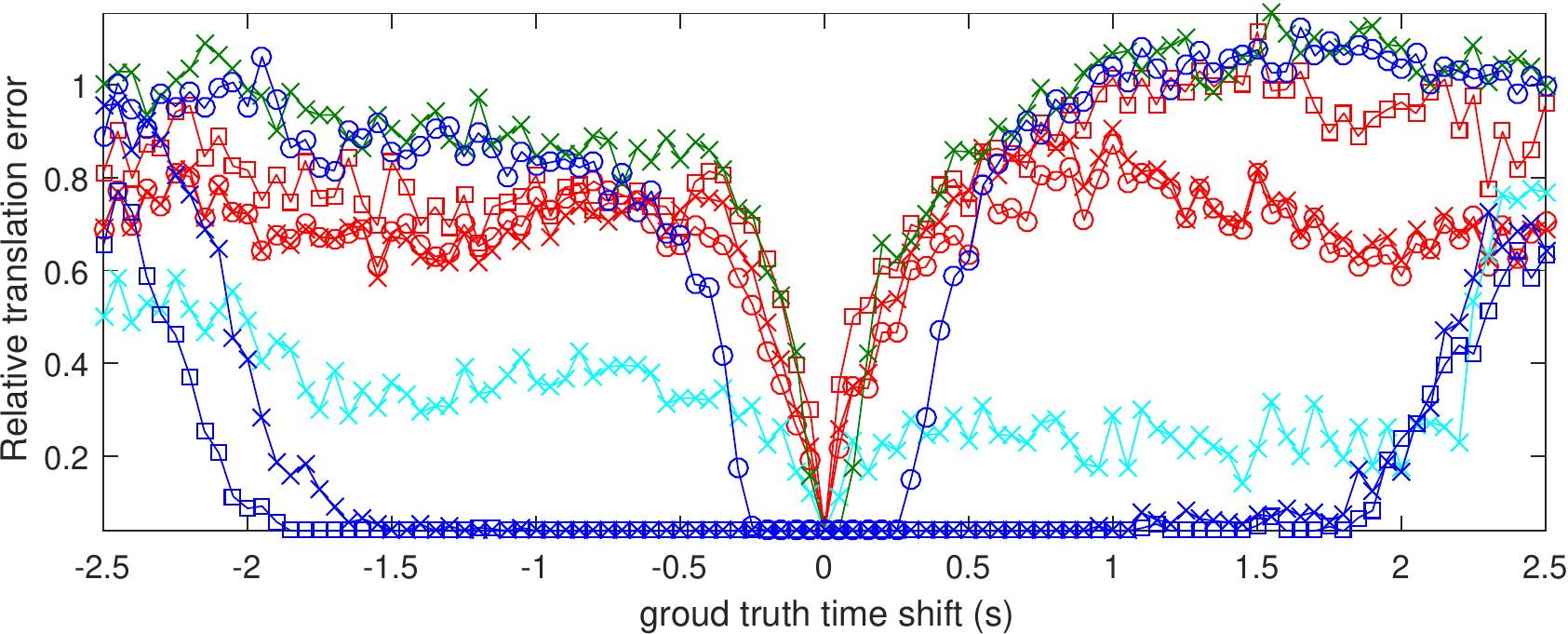}
    \includegraphics[width=\columnwidth]{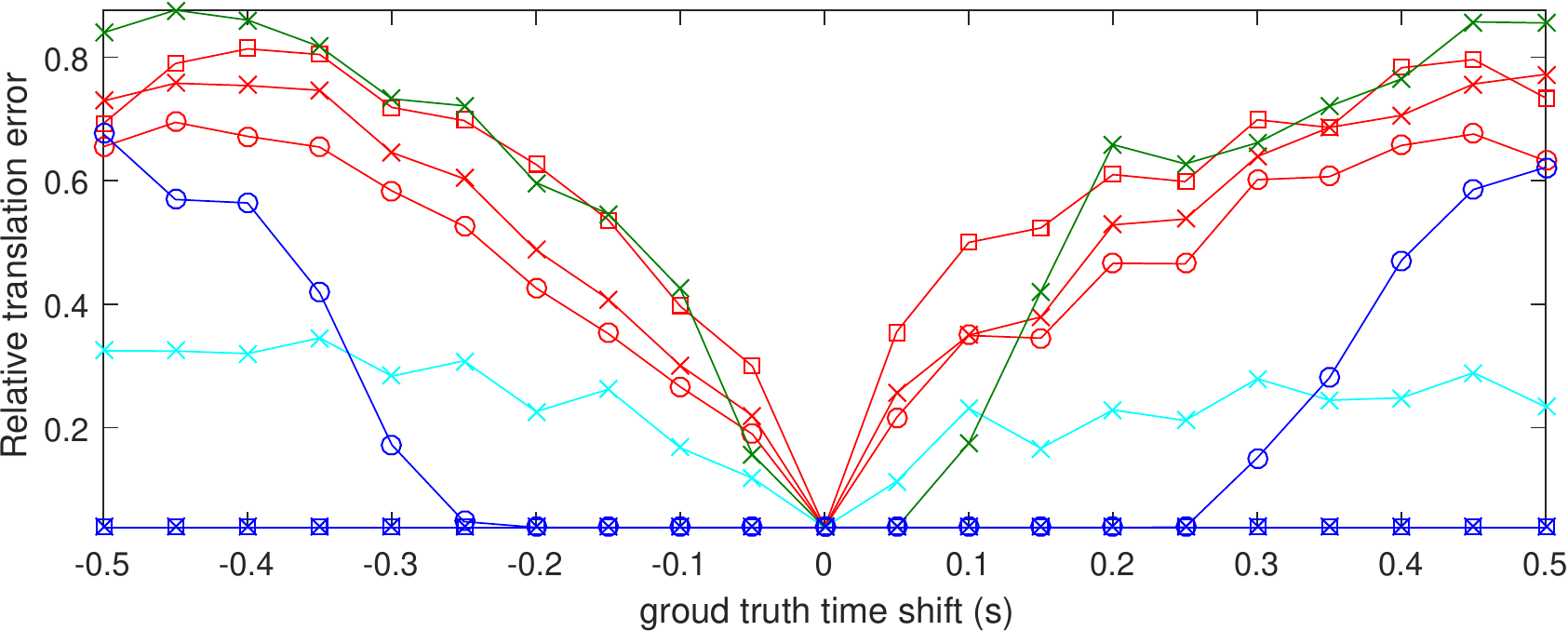}
    \caption{Error in relative rotation and translation between the two cameras from UvA dataset. All algorithms were tested, taking the resulting fundamental matrix and decomposing it into $\M{R}$ and $\V{t}$.}
    \label{fig:UvA_Rt}
\end{figure*}

The only real world dataset used for the main paper experiments for which the ground truth spatial calibration is provided is the UvA dataset. We extracted the ground truth relative $\M{R}_{gt}$ and $\V{t}_{gt}$ from the dataset camera matrices and compared them to the values estimated by all algorithms. Figure~\ref{fig:UvA_Rt} shows the angular error of $\M{R}$, measured as the rotation angle of $\M{R}_{err}=\M{R}^\top\M{R}_{gt}$, and the relative translation error measured as $||\V{t}_{gt}-\V{t}||$, where both $\V{t}_{gt}$ and $\V{t}$ are normalized to unit lengths. Errors are averaged over 100 runs for each datapoint.
The results follow the pattern of the results in Figure 4 of the paper, where when an algorithm successfully estimated the time shift, 
it also provided a good geometry estimate. 

Both iterative algorithms $T_\beta$-new-iter-pmax6 and $T_\beta$-new-iter-pmaxvar, which have $pmax$ large enough to cover the required time shifts, perform well over almost the entire range of time shifts. The efficient   $T_\beta$-new-iter-pmax0 which iteratively uses $d=1$ performed well up till the time shifts of 0.25~s (5 frames). $T_\beta$-new-d1, which is the solver using $d=1$ in RANSAC, was able to estimate the geometry reliably only for a time shift of 1 frame. All the algorithms based on the 7-point algorithm, including the 7-point algorithm itself in RANSAC, performed poorly on this dataset.
\section*{Acknowledgement}
\noindent This work was partly done during an internship of C. Albl and a postdoc position of Z. Kukelova at Microsoft Research Cambridge and was supported by the EU-H2020 project LADIO (number 731970), The Czech Science Foundation Project GACR P103/12/G084, Grant Agency of the CTU Prague projects SGS16/230/OHK3/3T/13, SGS17/185/OHK3/3T/13 and by SCCH GmbH under Project 830/8301544C000/13162.
\bibliographystyle{ieee}
\bibliography{synchronization}

\end{document}